\pdfoutput=1
\documentclass[11pt]{article}

\usepackage[final]{acl}

\usepackage{subcaption}
\usepackage{amsmath}
\usepackage{amssymb}
\usepackage{times}
\usepackage{latexsym}
\usepackage[normalem]{ulem}
\usepackage{algorithm}
\usepackage{algorithmic}
\usepackage[T1]{fontenc}
\usepackage[utf8]{inputenc}

\usepackage{microtype}

\usepackage{inconsolata}

\usepackage{graphicx}

\usepackage{multirow} 
\usepackage{booktabs}       % professional-quality tables
\usepackage{amsmath}        % for \text in math mode
\usepackage{amsfonts}       % blackboard math symbols
\title{
Sparse Subspace-to-Expert Sharing for Task-Agnostic Continual Learning}
\author{
\textbf{Fatema Siddika}\textsuperscript{1},
\textbf{Md Anwar Hossen}\textsuperscript{1},
\textbf{Tanwi Mallick}\textsuperscript{2},
\textbf{Ali Jannesari}\textsuperscript{1} \\
\textsuperscript{1}Iowa State University, Ames, USA \\
\textsuperscript{2}Argonne National Laboratory, USA \\
\texttt{\{fatemask, manwar, jannesar\}@iastate.edu} \\
\texttt{tmallick@anl.gov}
}
\begin{document}
\maketitle
\begin{abstract}
Continual learning in Large Language Models (LLMs) is hindered by the plasticity-stability dilemma, where acquiring new capabilities often leads to catastrophic forgetting of previous knowledge. Existing methods typically treat parameters uniformly, failing to distinguish between specific task knowledge and shared capabilities. We introduce Mixture of \textbf{S}parse \textbf{E}xperts for \textbf{T}ask \textbf{A}gnostic Continual Learning (SETA), a framework that resolves the plasticity-stability conflict through adaptive sparse subspace decomposition into task-specific expert modules. Unlike standard updates, where tasks compete for the same parameters, SETA separates knowledge into unique experts, designed to isolate task-specific patterns, and shared experts, responsible for capturing common features. This structure is maintained through adaptive elastic anchoring and a routing-aware regularization that jointly protect shared knowledge at both the weight and routing levels and enable a unified gating network to automatically retrieve the correct expert combination during inference. Extensive experiments across diverse domain-specific benchmarks demonstrate that SETA achieves competitive or superior overall performance relative to state-of-the-art continual learning baselines, with particularly strong retention of early-task knowledge and improved backward transfer on LLaMA-2 7B and Qwen3-4B. 
\end{abstract}
\section{Introduction}
Large Language Models have demonstrated remarkable performance across diverse natural language processing tasks, ranging from text classification and reasoning to dialogue and summarization~\cite{brown2020language, raffel2020exploring}. However, deploying LLMs in continual learning settings remains challenging~\cite{parisi2019continual} as models must adapt to sequential tasks while preserving previously acquired knowledge~\cite{kirkpatrick2017overcoming, li2017learning}. This challenge, known as catastrophic forgetting~\cite{li2017learning, kirkpatrick2017overcoming}, arises from the fundamental trade-off between learning plasticity, the rapid adaptation to new tasks, and memory stability, the retention of learned knowledge. The problem is further exacerbated in parameter-efficient fine-tuning regimes, where sparsity for efficiency increases susceptibility to overwriting historical information~\cite{han2015learning, evci2020rigging}.

Existing continual learning strategies generally fall into three primary categories, including Replay based methods that rehearse historical data~\cite{chaudhry2019tiny}, Regularization based techniques that penalize updates to critical parameters~\cite{kirkpatrick2017overcoming}, and Parameter Isolation approaches that physically separate task specific modules~\cite{rusu2016progressive}. Despite their differences, these approaches share a common limitation: they treat parameters uniformly within their respective subspaces~\cite{ren2024analyzing}. As a result, they fail to distinguish between generalizable features and task-specific requirements, leading to suboptimal trade-offs between stability and plasticity~\cite{li2025analyzing}.

To address these limitations, Continual Sparse Fine-Tuning provides a natural mechanism for uncovering the internal knowledge structure of large language models. Unlike parameter-efficient fine-tuning (PEFT)-based low-rank adapters that project updates through opaque dense matrices, sparse tuning identifies specific attention sub-blocks, particularly in the Value projection, based on high-utility gradient signals. This unique property serves as the foundation for our proposed framework, Mixture of Sparse Experts for Task Agnostic Continual Learning, referred to as \textbf{SETA}.
The design of SETA is driven by three fundamental research questions, each addressing a specific limitation in state-of-the-art literature:

\textit{\textbf{RQ1:} How can sparse parameter patterns be leveraged to decouple the acquisition of novel features from the retention of historical representations?}
To address RQ1, SETA introduces a subspace-based expert decomposition that frames sparse continual learning within a Mixture-of-Experts paradigm. By exploiting the inherent block-wise sparsity of gradients, the model is decomposed into distinct expert modules, ensuring that conflicting updates operate in isolated subspaces rather than competing for overlapping parameter blocks.

\textit{\textbf{RQ2:} Is it possible to automatically differentiate between reusable shared knowledge and task-specific unique knowledge within a single unified framework?}
To answer RQ2, the Split-on-Share (SoS) mechanism dynamically assigns 
parameters based on their recurrence across tasks. Overlapping parameters are designated as shared experts, protected through adaptive elastic anchoring and a routing-aware regularization that jointly prevent semantic drift without introducing additional hyperparameters, while non-overlapping parameters become unique experts that are strictly frozen as immutable memory to preserve historical knowledge.

\textit{\textbf{RQ3:} How can the relevant expert modules be dynamically 
retrieved during inference without relying on external task identifiers?}
SETA resolves RQ3 by implementing adaptive gating expansion with logit 
invariance, which preserves existing routing decision boundaries during expert splitting, and a routing-aware regularization that shapes the gate's routing decisions during training to reduce interference with shared experts. This 
allows the router to evolve alongside the experts and dynamically select 
relevant parameters based solely on input tokens, enabling task-agnostic inference without explicit task labels.

Section 2 formulates the problem, Section 3 introduces SETA and the Split-on-Share mechanism, Section 4 presents experiments, Section 5 related works and Section 6 concludes.
\section{Problem Formulation and Motivation}
\subsection{Challenges of Continual Learning}
Continual learning aims to enable models to acquire new capabilities 
from a sequence of tasks $\mathcal{T} = \{T_1, \dots, T_k\}$ without 
forgetting previously learned knowledge~\cite{wu2024continual}, creating 
a fundamental plasticity-stability trade-off. Full fine-tuning offers high 
plasticity but is computationally expensive and disrupts the optimization 
landscape, leading to catastrophic forgetting~\cite{luo2025empirical}. 
To address this, parameter-efficient fine-tuning (PEFT) methods such as 
LoRA freeze the backbone and train small low-rank modules~\cite{hu2022lora}. 
However, restricting updates to a narrow bottleneck dimension ($r \ll d$) 
creates a geometric disconnect between sequential task minima~\cite{ren2024analyzing}, 
introducing noise that hinders knowledge retention over long task 
sequences and causes a plasticity-stability failure where new learning 
overwrites historical parameters.

\begin{figure*}[htbp]
    \centering
    \includegraphics[width=1.02\linewidth]{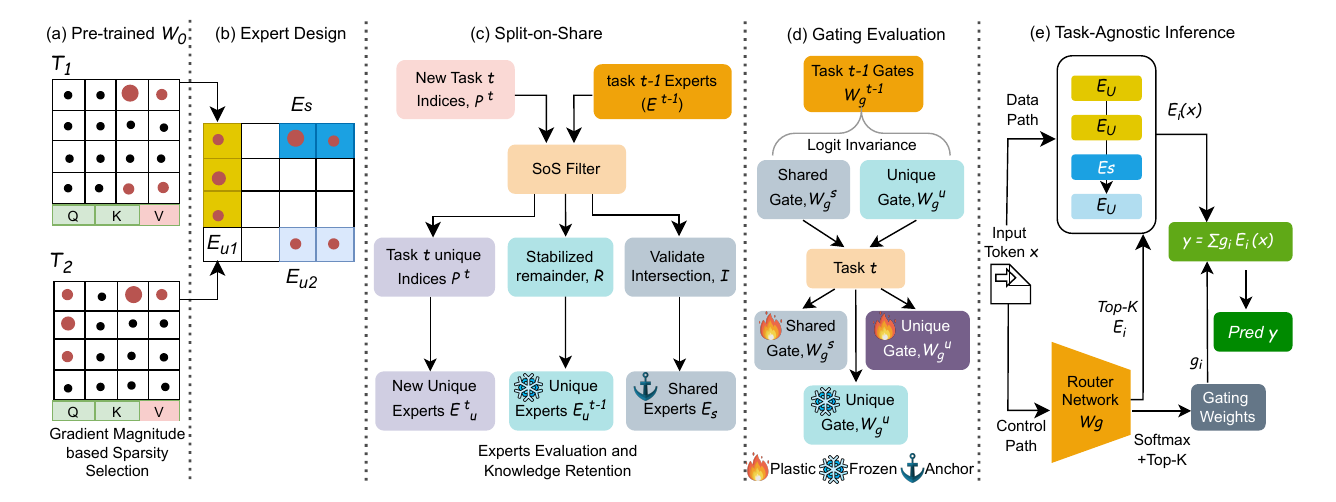}
    % \caption{The SETA system utilizes a specialized sparse fine-tuning pipeline where task-specific parameters are isolated through a Split-on-Share (SoS) mechanism. This structure partitions the model into Shared $E_{\text{s}}$ and Unique $E_{\text{u}}$ experts to resolve parameter collisions while employing a task-agnostic gating network.}
    \caption{\textbf{Overview of the SETA Framework Architecture.} 
\textbf{(a--b) Sparse Subspace Selection:} High-utility parameter blocks are identified from the pre-trained LLM using gradient magnitude to form the expert design. 
\textbf{(c) Split-on-Share (SoS) Evolution:} The SoS filter partitions parameters into plastic Shared ($E_{\text{s}}$) and frozen Unique ($E_{\text{u}}$) experts to resolve parameter collisions and retain knowledge. 
\textbf{(d) Gating Evaluation:} The gating network expands using logit invariance to strictly preserve decision boundaries during expert splitting. 
\textbf{(e) Task-Agnostic Inference:} A router network dynamically weights all experts via softmax for input tokens, enabling automatic task processing without task identifications.}
\label{fig:seta_architecture}
    \label{fig:seta_diagram}
\end{figure*}
\subsection{Shared Subspace Dilemma}
The core failure of continual learning arises from the Shared Subspace Dilemma. As tasks are learned sequentially, parameters often become important for multiple tasks. Let $\mathcal{P}_t$ denote parameters for task $t$; its intersection with the cumulative historical subspace $\mathcal{P}_{1:t-1} = \bigcup_{k=1}^{t-1} \mathcal{P}_k$ defines a shared region $\mathcal{I} = \mathcal{P}_t \cap \mathcal{P}_{1:t-1}$. Here, updating weights risks overwriting prior knowledge, while freezing limits plasticity. Standard methods address this via uniform regularization~\cite{kirkpatrick2017overcoming}:
\begin{equation}
    \mathcal{L}_{\text{reg, uniform}} = \gamma \cdot \| W - W^{*}_{1:t-1} \|^2,
\end{equation}
where $\gamma$ is a fixed penalty. However, uniform regularization is too rigid, treating parameters equally rather than distinguishing shared features requiring controlled plasticity from task-specific regions needing stability~\cite{lopez2017gradient}. Since static penalties cannot simultaneously prevent semantic drift and 
enable new learning~\cite{farajtabar2020orthogonal}, resolving this 
trade-off requires both a structural solution to isolate parameter subspaces 
and an adaptive regularization strategy that responds to each expert's 
accumulated task history and live weight displacement.
\subsection{Mixture of Sparse Experts for CL}
We propose SETA, a modular framework for sparse subspace adaptation. SETA resolves the stability–plasticity conflict by adaptively decomposing task-specific sparse subspaces into unique and shared experts, preventing parameter overwriting. When subspaces overlap beyond a layer-wise significance threshold, the Split-on-Share mechanism decouples contested regions, enabling stable structural evolution. Combined with adaptive elastic anchoring, routing-aware regularization, and logit-invariant gating expansion, this design supports cumulative learning and task-agnostic inference without explicit task identifiers.

\section{Methodology} 
% This section presents the SETA methodology. A detailed algorithm is provided in the appendix \ref{app:algorithm}.
\subsection{Sparse Subspace Selection}
\label{sec:sparse_construction}
We aim to fine-tune only the most task-relevant sparse sub-matrices, 
defining the subspace as a block-sparse optimization boundary through 
gradient-driven selection, rather than a conceptually disentangled 
semantic latent space.
We partition pre-trained weight matrices $W \in \mathbb{R}^{d \times m}$ 
into a grid of fixed-size sub-blocks $\mathbb{B} = \{B_{i,j}\}$ of 
dimension $l \times l$, where $l$ is determined per architecture as the 
greatest common divisor of the model's matrix dimensions. This 
architecture-aware sizing prevents remainder mismatches, ensures 
contiguous memory alignment, and maximizes hardware utilization 
across all supported model families. We identify high-utility regions via a parameter-free scoring 
pipeline executed during a warm-up phase before each task.\\
\textbf{Dynamic Warm-up Stopping:}
We monitor the per-step gradient contribution ratio:
Training halts when $\rho(t)$ falls below its own coefficient of 
variation $\text{CV} = \sigma_\rho / \mu_\rho$ over a sliding window 
of $W=10$ steps, adapting warm-up length to each task's gradient dynamics.
\begin{equation}
    \rho(t) = \frac{\|\nabla W\|_t}{\sum_{s \leq t} \|\nabla W\|_s}.
\end{equation}
\textbf{Layer-Normalised Scoring:}
The raw block importance score is computed as the mean absolute gradient 
within each block:
\begin{equation}
    \mathcal{S}_{i,j}^{(k)} = \frac{1}{l^2} \sum_{(u,v) \in B_{i,j}} |\nabla W_{u,v}|
\end{equation}
where $\nabla W_{u,v}$ indicates parameter sensitivity.
Since gradient magnitudes vary across layers, each layer's score matrix 
is standardized to zero mean and unit variance before global ranking, 
preventing dominant layers from monopolizing the block selection. \\
\textbf{Cross-Task Divergence Scoring:}
For tasks $t > 1$, the normalized score is augmented with a divergence 
signal rewarding blocks with novel gradient response relative to prior 
tasks. The final score is a convex combination:
\begin{equation}
    \tilde{\mathcal{S}}_t = \alpha \cdot \hat{\mathcal{S}}_t + 
    (1-\alpha) \cdot \left|\hat{\mathcal{S}}_t - \bar{\mathcal{S}}_{1:t-1}\right|,
\end{equation}
where $\bar{\mathcal{S}}_{1:t-1}$ is the mean block score over all prior tasks 
and $\alpha = 1/(1 + \log t)$ decays automatically with task count, introducing 
no additional hyperparameter.

\textbf{Adaptive Block Budget:}
The block budget is parameterised as a fraction $\rho$ of total available 
blocks, analogous to a LoRA rank, giving $N_{\text{base}} = \lfloor \rho \cdot 
|\mathbb{B}| \rfloor$, with per-layer quota allocated inversely proportional 
to historical coverage. Selection proceeds beyond $N_{\text{base}}$ until the 
marginal rate of previously unselected blocks stabilizes, allowing later tasks 
to naturally expand their budget without manual adjustment.

As shown in Figure~\ref{fig:gradient_analysis} (Appendix~\ref{app:phase0:overlap}), $\sim$95\% of 
high-magnitude gradients concentrate in the Value ($V$) projection, 
attributed to Softmax Saturation where $Q$ and $K$ gradients vanish as 
$QK^\top / \sqrt{d_k}$ grows. We therefore restrict block selection to the 
V-projection. Since subspaces inevitably overlap across tasks, we 
consolidate common features into shared experts while isolating 
task-specific regions, mitigating catastrophic forgetting.

\subsection{Design of Unique and Shared Experts}
We partition the sparse subspace $\mathcal{P}_t$ into two functional 
sets based on its overlap with the cumulative historical subspace 
$\mathcal{P}_{1:t-1}$. The task-specific unique experts $E_{\text{u}}$ are constructed from the disjoint sparse subspace defined by indices $\mathcal{P}_t \setminus \mathcal{P}_{1:t-1}$. These parameters are exclusive to $T_t$.
In contrast, the shared experts $E_{\text{s}}$ occupy the intersection subspace defined by $\mathcal{P}_t \cap \mathcal{P}_{1:t-1}$. These parameters represent the structural feature overlap between tasks, serving as a basis for positive backward transfer. The effective sparse subspace $W$ is thus formalized as the summation of the frozen base model $W_0$ and the functional deltas:
\begin{equation}
    W = W_0 + \sum_{j \in \mathcal{E}_{\text{u}}} \Delta W_j + \sum_{i \in \mathcal{E}_{\text{s}}} \Delta W_i.
\end{equation}
\subsection{Split-on-Share (SoS) Expert Evolution}
\label{sec:split_on_share}
In continual learning, the emergence of overlapping sub-blocks between $\mathcal{P}_{t-1}$ and $\mathcal{P}_{t}$ necessitates structural evolution. We propose the Robust Split-on-Share (SoS) algorithm, which governs this process via a layer-wise topological filter designed to distinguish semantic commonality from stochastic noise. For each layer $l$, we first quantify the raw intersection $\mathcal{I}_l = \mathcal{P}_{t-1}^{(l)} \cap \mathcal{P}_{t}^{(l)}$ and the raw unique remainder $\mathcal{R}_l = \mathcal{P}_{t-1}^{(l)} \setminus \mathcal{P}_{t}^{(l)}$. To ensure robust modularity, we apply a two-stage conditional set operation using the Expert Creation Threshold ($\tau_{\text{ect}}$) and the Tiny Remainder Threshold ($\tau_{\text{trt}}$):

\textbf{Design choice of Expert Creation and Tiny Remainder Threshold: } 
The Expert Creation Threshold ($\tau_{\text{ect}}$) and Tiny Remainder 
Threshold ($\tau_{\text{trt}}$) are fixed topological thresholds that 
filter structural noise rather than scaling continuous penalties. Their primary purpose is to distinguish stable semantic commonality from stochastic gradient noise. Specifically, $\tau_{\text{ect}}$ prevents the formation of diluted experts by rejecting parameter intersections that lack sufficient cardinality to represent a meaningful functional unit. Conversely, $\tau_{\text{trt}}$ mitigates architectural fragmentation by absorbing negligible unique residuals into the shared manifold, thereby preserving computational efficiency and gating stability. \\
\textbf{Filtering Small Overlaps:}
We validate if the intersection represents meaningful shared knowledge. If the overlap cardinality $|\mathcal{I}_l|$ is below $\tau_{\text{ect}}$, we reject it as a stochastic coincidence and reintegrate the indices into the unique subspace:
\begin{equation}
\label{eq:sos_Filtering_Small_Overlaps}
    \text{if } |\mathcal{I}_l| < \tau_{\text{ect}} \implies \mathcal{R}_l \leftarrow \mathcal{R}_l \cup \mathcal{I}_l, \quad \mathcal{I}_l \leftarrow \emptyset.
\end{equation}
\textbf{Merging Tiny Fragments:}
We analyze the sparsity of the remaining unique indices to prevent fragmentation. If the residual $|\mathcal{R}_l|$ falls below $\tau_{\text{trt}}$, it indicates a negligible specialized structure. These indices are absorbed into the shared manifold to preserve computational efficiency:
\begin{equation}
\label{eq:sos_Merging_Tiny_Fragments}
    \text{if } 0 < |\mathcal{R}_l| < \tau_{\text{trt}} \implies \mathcal{I}_l \leftarrow \mathcal{I}_l \cup \mathcal{R}_l, \quad \mathcal{R}_l \leftarrow \emptyset.
\end{equation}
\textbf{Expert Formation:}
Based on the filtered sets, the architecture executes the split. The validated intersection $\mathcal{I}_l$ instantiates the plastic shared expert $E_{\text{s}}$, utilizing weight inheritance $W_{\text{s}} \leftarrow W_{t-1}[\mathcal{I}_l]$. The stabilized residuals $\mathcal{R}_l$ form the frozen unique expert $E_{\text{u}}^{(t-1)}$. Finally, indices exclusive to the new domain $\mathcal{P}_{t}^{(l)} \setminus (\mathcal{I}_l \cup \mathcal{R}_l)$ initialize the New Task Expert $E_{\text{u}}^{(t)}$. This mechanism enables sub-linear capacity expansion by selectively merging redundant functionalities. Empirically, we observe that this growth follows a logarithmic trend; however, we note that this scaling behavior is a heuristic observation rather than a strictly proven theoretical bound.

\subsection{Catastrophic Forgetting Prevention}
\label{sec:knowledge_preservation}
Catastrophic forgetting is prevented through a dual strategy of subspace isolation to protect history and adaptive regularization to stabilize shared knowledge.

\textbf{Orthogonal Subspace Isolation:}
We strictly divide the sparse subspace to isolate and protect historical memories. During the training of task $t$, all prior unique experts $E_{\text{u}}^{(k)}$ ($k < t$) are frozen, which is enforced via a zero-gradient constraint:
\begin{equation}
    \forall w \in \bigcup_{k < t} \mathcal{E}_{\text{u}}^{(k)}, \quad \frac{\partial \mathcal{L}}{\partial w} \equiv 0.
\end{equation}
Optimization is restricted to the active topology: the plastic shared expert $E_{\text{s}}$ and the current unique expert $E_{\text{u}}^{(t)}$. This ensures that features specific to past domains remain immutable.

\textbf{Adaptive Elastic Anchoring:}
While the Shared Expert $E_{\text{s}}$ facilitates forward transfer, it is 
vulnerable to semantic drift. We mitigate this with a per-expert dynamic 
regularisation coefficient that adapts at every training step based on two 
signals: the accumulated task history of the expert and its live weight 
displacement.

Let $\hat{W}_i$ be the snapshot of shared expert $i$ taken before task $k$ 
begins. The accumulation weight $\omega_i = n_{\text{prior}}^{(i)} / (k-1)$ 
scales protection proportionally to the number of prior task IDs encoded in 
expert $i$, while the drift factor $\delta_i(t)$ monitors live weight 
displacement relative to the pre-task snapshot:
\begin{equation}
    \delta_i(t) = 1 + \tanh\!\left(\frac{\|W_i^{(t)} - \hat{W}_i\|}
    {\max(\|\hat{W}_i\|,\,\epsilon)}\right),
\end{equation}
yielding the per-expert dynamic regularisation coefficient:
\begin{equation}
    \lambda_i(t) = \lambda_{\text{base}} \cdot \omega_i \cdot \delta_i(t),
\end{equation}
where $n_{\text{prior}}^{(i)}$ is the number of prior task IDs encoded in 
expert $i$. The drift factor $\delta_i(t) \in [1, 2)$ is non-decreasing as the expert drifts ($\delta_i=1$ when $W_i^{(t)}=\hat{W}_i$), self-correctively strengthening the penalty in proportion to the displacement magnitude. The regularization loss is:
\begin{equation}
   \mathcal{L}_{\text{reg}} = \sum_{i \in \mathcal{E}_{\text{s}}} 
    \lambda_i(t) \cdot \|W_i^{(t)} - \hat{W}_i\|^2.
\end{equation}

\subsection{Adaptive Gating Expansion}
To ensure the model remains stable when splitting the historical expert $E_{t-1}$ into separate experts, we strictly preserve the pre-activation logits.\\
\textbf{Logit Invariance via Weight Inheritance:}
Let the gating parameter for the active expert at task $t-1$ be denoted as $W_g^{(t-1)}$. For an input token $\mathbf{x}$, the routing logit is defined as $z = \mathbf{x}^\top W_g^{(t-1)}$. 

Upon executing the experts split operation, we initialize the gate for the plastic shared expert $W_g^{(\text{s})}$ and the frozen task-specific Expert $W_g^{(\text{u})}$. This approximately preserves the routing logits for both successors, with a small perturbation added to $W_g^{(\text{u})}$ to break symmetry. By keeping the original weight geometry, we preserve the router's existing decision boundaries, preventing cold-start errors after the split.

\subsection{Routing-Aware Gating Regularization}
\label{sec:gating_rl}
While $\mathcal{L}_{\text{reg}}$ acts on expert weights after 
gradient updates have occurred, we additionally shape the gating 
network's routing decisions before drift accumulates by 
penalizing high routing probability toward at-risk shared experts:
\begin{equation}
    \mathcal{L}_{\text{gate}} = \sum_{i \in \mathcal{E}_{\text{s}}} 
    \lambda_i \cdot \mathbb{E}_{x}\!\left[\sigma(g_\theta(x))_i\right],
\end{equation}
where $\sigma(g_\theta(x))_i$ is the softmax routing probability to 
expert $i$ and $\lambda_i$ is the same coefficient as in $\mathcal{L}_{\text{reg}}$, 
introducing no additional hyperparameter.
Together, $\mathcal{L}_{\text{reg}}$ and $\mathcal{L}_{\text{gate}}$ 
form a two-layer defence: the gate is steered away from risky routing 
decisions before drift accumulates, while weight regularization catches 
any residual drift.
\subsection{Objective Function}
For the initial task $T_1$, we minimize the standard sparse fine-tuning loss. For subsequent tasks ($n > 1$), we optimize the active subspace of the new unique expert $E_{\text{u}}^{(t)}$ and the shared expert $E_{\text{s}}$ while the Base Model $W_0$ and past unique experts remain frozen. The composite objective is:

\begin{equation}
\label{equ:seta_train_object}
    \mathcal{L} = -\sum_{(x,y)\sim \mathcal{D}_t} \log P(y \mid x; \Theta_t) 
    + \mathcal{L}_{\text{reg}} + \mathcal{L}_{\text{gate}},
\end{equation}
where $\Theta_t = W_0 + \sum_{k \in \mathcal{E}_{\text{u}}^{(t)}} \Delta W_k + 
\sum_{j \in \mathcal{E}_{\text{s}}} \Delta W_j$ denotes active parameter 
configuration at task $t$.
\begin{table*}[t]
\centering
\caption{Continual learning performance across a six-task sequence using LLaMA-2 7B, and Qwen3-4B. Performance is reported as the Overall Performance (OP), Backward Transfer (BWT), and Average Forgetting ($F_T$) after sequential training. Higher OP ($\uparrow$) and BWT ($\uparrow$) indicate better final task accuracy and less negative transfer; lower $F_T$ ($\downarrow$) indicates less catastrophic forgetting of previously learned tasks. \textbf{Bold} values indicate the best performance in each column, \underline{underlined} values denote the second-best.}
\label{tab:domain_specific_results}
\resizebox{\textwidth}{!}{%
\begin{tabular}{lcccccccc c}
\toprule
 & \multicolumn{6}{c}{\textbf{LLaMA-2 7B: Continual Learning on TRACE Benchmarks}} 
 & \multicolumn{3}{c}{\textbf{Metrics}} \\
\cmidrule(lr){2-7} \cmidrule(lr){8-10}
Method & C-STANCE & FOMC & MeetingBank & ScienceQA & NumGLUE-cm & 20Minuten 
       & OP $\uparrow$ & BWT $\uparrow$ & $F_T$ $\downarrow$ \\
\toprule
Full FT      & 49.60 & 62.10 & 58.03 & 61.10 & 35.58 & 50.28 & 52.78 & -- & -- \\
\midrule
Seq-LoRA & 0.2  &  1.2  &  8.7  & 15.8  & 14.8  & 41.4  & 13.68 & -31.24 & 31.24 \\
EWC       & 19.0  & 24.8  & 15.2  & 42.4  & 14.8  & 41.50 & 26.28 & -22.96 & 22.96 \\
GEM       &  5.2  &  7.1  & 12.0  & 24.0  & 17.3  & 40.9  & 17.75 & -26.10 & 26.10 \\
I-LoRA    & 33.6  &  8.1  & 14.4  & 44.2  & 22.2  & 41.4  
          & \underline{27.32} & \underline{-18.30} & \textbf{18.30} \\
\midrule
SETA (ours) & 27.31 & 26.41 & 20.81 & 41.63 & 22.55 & 33.62 
            & \textbf{28.72} & \textbf{-16.33} & \underline{19.10} \\
\toprule
 & \multicolumn{6}{c}{\textbf{Qwen3-4B: Continual Learning on TRACE Benchmarks}}
 & \multicolumn{3}{c}{\textbf{Metrics}} \\
\cmidrule(lr){2-7} \cmidrule(lr){8-10}
Method & C-STANCE & FOMC & MeetingBank & ScienceQA & NumGLUE-cm & 20Minuten
       & OP $\uparrow$ & BWT $\uparrow$ & $F_T$ $\downarrow$ \\
\midrule
SeqLoRA &  0.2  & 21.5  &  5.2  & 33.3  & 51.9  & 38.9  & 25.17 & -35.54 & 35.54 \\
EWC     &  0.2  &  0.6  &  4.2  &  6.4  & 46.9  & 38.6  & 16.15 & -46.54 & 46.54 \\
GEM     &  0.5  & 13.5  &  6.5  & 67.7  & 29.6  & 38.6  & 26.07 & -32.70 & 32.70 \\
I-LoRA  & 27.2  & 38.5  & 16.0  & 71.0  & 37.0  & 39.2  
        & \underline{38.15} & \underline{-15.54} & \underline{15.54} \\
\midrule
SETA (ours) & 63.64 & 33.0 & 21.42 & 80.77 & 25.97 & 34.98 
            & \textbf{43.30} & \textbf{-12.67} & \textbf{15.42} \\
\bottomrule
\end{tabular}%
}
\end{table*}

\subsection{Task-Agnostic Inference}
SETA achieves task-agnostic deployment via a unified content-based 
routing network. The gating network acts as a unified linear transformation $W_g \in \mathbb{R}^{N \times d}$, mapping input $\mathbf{x}$ to a global logit vector over the expert population $N = |\mathcal{E}_{\text{s}}| + |\mathcal{E}_{\text{u}}|$. This single-layer architecture enables simultaneous evaluation, computing the output as a weighted superposition of the frozen backbone $W_0$ and retrieved sparse functional deltas:
\begin{equation}
\label{eq:task_agnostic_pred}
\begin{split}
    \mathbf{y} = W_0 \mathbf{x} + \sum_{k \in \mathcal{E}_{\text{s}}} \sigma_k(\mathbf{x} W_g^\top) \Delta W_k \mathbf{x} \\
    + \sum_{k \in \mathcal{E}_{\text{u}}} \sigma_k(\mathbf{x} W_g^\top) \Delta W_k \mathbf{x},
\end{split}
\end{equation}
where, $\sigma(\cdot)$ denotes the softmax operator. If a token shares features with multiple domains, $W_g$ automatically activates multiple experts without requiring discrete task IDs.
\section{Experiments}
\label{sec:experiments}
\subsection{Evaluation Benchmarks}
To rigorously evaluate LLM plasticity and retention, we use two benchmark 
categories spanning domain specificity, format diversity, and 
reasoning complexity. The Continual Learning Benchmarks include 
ScienceQA for education, FOMC for finance, MeetingBank for political 
discourse, C-STANCE and 20Minuten for multilingual adaptation, and 
NumGLUE for mathematical reasoning. The expert learning rate is $1e$-4 with a 0.2 linear warmup. We sample 5,000 training and 500 evaluation instances per dataset.
\subsection{Evaluation Metrics}
Let $M_{i,j}$ denote the inference accuracy on task $j$ after training 
on task $i$. We evaluate using three complementary metrics. Overall 
Performance ($\mathrm{OP}$) measures the final model's average accuracy 
across all tasks:
\begin{equation}
    \mathrm{OP} = \frac{1}{T} \sum_{j=1}^{T} M_{T,j}.
\end{equation}
Average Forgetting ($F_{T}$) quantifies the decline from each task's 
peak accuracy:
\begin{equation}
    F_{T} = \frac{1}{T-1} \sum_{j=1}^{T-1} \max_{l \leq T} (M_{l,j} - M_{T,j}).
\end{equation}
Backward Transfer ($\mathrm{BWT}$) measures the influence of new tasks 
on previously learned ones:
\begin{equation}
    \mathrm{BWT} = \frac{1}{T-1} \sum_{j=1}^{T-1} (M_{T,j} - M_{j,j}).
\end{equation}
Higher $\mathrm{OP}$ $\uparrow$ and $\mathrm{BWT}$ $\uparrow$, and 
lower $F_{T}$ $\downarrow$ indicate better continual learning performance.

\subsection{Comparative Analysis of CL}

\textbf{Evaluation of Domain Adaptation:}
We evaluated LLaMA-2 7B and Qwen3-4B models using average inference
accuracy $Acc_t$ as the primary metric. Table~\ref{tab:domain_specific_results}
confirms that SETA establishes a new state of the art, outperforming
strong baselines such as I-LoRA on most tasks. On LLaMA-2 7B,
SETA demonstrates stronger initial task acquisition and substantially
greater resistance to catastrophic collapse: while I-LoRA degrades to
8.1\% on FOMC by the final step, SETA retains 26.41\%, a gap of over
18 percentage points. This robustness extends to MeetingBank, where
SETA preserves 20.81\% versus 14.4\% for I-LoRA, confirming that
sparse gradient-driven adaptation captures specialized knowledge more
reliably than replay-based methods. \\
\textbf{Overall Performance Score and Forgetting:}
Table~\ref{tab:domain_specific_results} compares SETA with established
baselines in the overall performance score $OP$ and forgetting $F_T$.
SETA achieves an $OP$ of 28.72\% on LLaMA-2 7B, exceeding I-LoRA at
27.32\%, while its $F_T$ of 19.10\% remains competitive against
I-LoRA's 18.30\%. On Qwen3-4B, SETA achieves the best $OP$ of 43.30\%
and the lowest $F_T$ of 15.42\%, outperforming I-LoRA on both metrics.
Taken together, these results confirm that the sparse update mechanism
balances stability and plasticity without compromising final utility.
An ablation isolating the effect of $\mathcal{L}_{\text{gate}}$ is 
provided in Appendix~\ref{app:ablation:gating}.
\begin{table*}[htbp]
    \centering
    \caption{Task-wise inference accuracy of continual learning methods 
    on Qwen3-4B. Each row corresponds to the model state after training 
    on that task, evaluated on all tasks seen so far. Tasks are ordered 
    sequentially: T1=C-STANCE, T2=FOMC, T3=MeetingBank, T4=ScienceQA, 
    T5=NumGLUE-cm, T6=20Minuten. Diagonal entries reflect plasticity; 
    off-diagonal decay reflects forgetting.}
    \label{tbl:task_wise_performance_qwen3_4b}
    \renewcommand{\arraystretch}{1.3}
    \setlength{\tabcolsep}{3pt}
    \resizebox{\textwidth}{!}{
        % Table 1: EWC
        \begin{tabular}{r |c|c|c|c|c|c|}
            \multicolumn{1}{c}{} & \multicolumn{6}{c}{\textbf{EWC}} \\ \cline{1-7}
            \multicolumn{1}{c}{\scriptsize Train $\setminus$ Test} & \multicolumn{1}{c}{T1} & \multicolumn{1}{c}{T2} & \multicolumn{1}{c}{T3} & \multicolumn{1}{c}{T4} & \multicolumn{1}{c}{T5} & \multicolumn{1}{c}{T6} \\ \hline
            T1 & 59.8 & \multicolumn{5}{c}{} \\ \cline{2-2} \cline{3-3}
            T2 & 50.4 & 67.3 & \multicolumn{4}{c}{} \\ \cline{2-3} \cline{4-4}
            T3 & 0.2  & 42.3 & 19.2 & \multicolumn{3}{c}{} \\ \cline{2-4} \cline{5-5}
            T4 & 5.2  & 36.1 & 22.0 & 84.2 & \multicolumn{2}{c}{} \\ \cline{2-5} \cline{6-6}
            T5 & 5.8  & 52.2 & 17.0 & 75.2 & 60.5 & \multicolumn{1}{c}{} \\ \cline{2-6} \cline{7-7}
            T6 & 0.2  & 0.6  & 4.2  & 6.4  & 46.9 & 38.6 \\ \cline{2-7}
        \end{tabular}
        \quad
        % Table 2: I-LoRA
        \begin{tabular}{r |c|c|c|c|c|c|}
            \multicolumn{1}{c}{} & \multicolumn{6}{c}{\textbf{I-LoRA}} \\ \cline{1-7}
            \multicolumn{1}{c}{\scriptsize Train $\setminus$ Test} & \multicolumn{1}{c}{T1} & \multicolumn{1}{c}{T2} & \multicolumn{1}{c}{T3} & \multicolumn{1}{c}{T4} & \multicolumn{1}{c}{T5} & \multicolumn{1}{c}{T6} \\ \hline
            T1 & 59.0 & \multicolumn{5}{c}{} \\ \cline{2-2} \cline{3-3}
            T2 & 50.6 & 66.3 & \multicolumn{4}{c}{} \\ \cline{2-3} \cline{4-4}
            T3 & 15.8 & 59.5 & 19.0 & \multicolumn{3}{c}{} \\ \cline{2-4} \cline{5-5}
            T4 & 33.0 & 45.8 & 16.4 & 82.4 & \multicolumn{2}{c}{} \\ \cline{2-5} \cline{6-6}
            T5 & 19.8 & 45.4 & 18.4 & 80.6 & 40.7 & \multicolumn{1}{c}{} \\ \cline{2-6} \cline{7-7}
            T6 & 27.2 & 38.5 & 16.0 & 71.0 & 37.0 & 39.2 \\ \cline{2-7}
        \end{tabular}
        \quad
        % Table 3: SETA
        \begin{tabular}{r |c|c|c|c|c|c|}
            \multicolumn{1}{c}{} & \multicolumn{6}{c}{\textbf{SETA (ours)}} \\ \cline{1-7}
            \multicolumn{1}{c}{\scriptsize Train $\setminus$ Test} & \multicolumn{1}{c}{T1} & \multicolumn{1}{c}{T2} & \multicolumn{1}{c}{T3} & \multicolumn{1}{c}{T4} & \multicolumn{1}{c}{T5} & \multicolumn{1}{c}{T6} \\ \hline
            T1 & 57.01 & \multicolumn{5}{c}{} \\ \cline{2-2} \cline{3-3}
            T2 & 57.99 & 67.54 & \multicolumn{4}{c}{} \\ \cline{2-3} \cline{4-4}
            T3 & 52.12 & 56.45 & 46.69 & \multicolumn{3}{c}{} \\ \cline{2-4} \cline{5-5}
            T4 & 37.29 & 26.00 & 24.16 & 75.90 & \multicolumn{2}{c}{} \\ \cline{2-5} \cline{6-6}
            T5 & 59.00 & 38.91 & 22.83 & 83.01 & 41.03 & \multicolumn{1}{c}{} \\ \cline{2-6} \cline{7-7}
            T6 & 63.64 & 33.00 & 21.42 & 80.77 & 25.97 & 34.98 \\ \cline{2-7}
        \end{tabular}
    }
\end{table*}
\begin{table*}[htbp]
    \centering
    \caption{Task-wise inference accuracy of continual learning methods 
on LLaMA-2 7B. Each row corresponds to the model state after training 
on that task, evaluated on all tasks seen so far. Diagonal entries 
reflect plasticity on the current task; off-diagonal decay reflects 
forgetting of prior tasks.}
    \label{tbl:task_wise_performance}
    \renewcommand{\arraystretch}{1.3}
    \setlength{\tabcolsep}{3pt}
    \resizebox{\textwidth}{!}{
        % Table 1: PP
        \begin{tabular}{r |c|c|c|c|c|c|}
            \multicolumn{1}{c}{} & \multicolumn{6}{c}{\textbf{PP}} \\ \cline{1-7}
            \multicolumn{1}{c}{\scriptsize Train $\setminus$ Test} & \multicolumn{1}{c}{T1} & \multicolumn{1}{c}{T2} & \multicolumn{1}{c}{T3} & \multicolumn{1}{c}{T4} & \multicolumn{1}{c}{T5} & \multicolumn{1}{c}{T6} \\ \hline
            T1 & 37.2 & \multicolumn{5}{c}{} \\ \cline{2-2} \cline{3-3}
            T2 & 32.4 & 51.8 & \multicolumn{4}{c}{} \\ \cline{2-3} \cline{4-4}
            T3 & 33.2 & 25.4 & 21.1 & \multicolumn{3}{c}{} \\ \cline{2-4} \cline{5-5}
            T4 & 32.4 & 24.2 & 14.7 & 44.2 & \multicolumn{2}{c}{} \\ \cline{2-5} \cline{6-6}
            T5 & 16.4 & 49.0 & 7.9 & 26.2 & 27.2 & \multicolumn{1}{c}{} \\ \cline{2-6} \cline{7-7}
            T6 & 32.0 & 13.1 & 9.3 & 40.8 & 19.8 & 40.6 \\ \cline{2-7}
        \end{tabular}
        \quad
        % Table 2: I-LoRA
        \begin{tabular}{r |c|c|c|c|c|c|}
            \multicolumn{1}{c}{} & \multicolumn{6}{c}{\textbf{I-LoRA}} \\ \cline{1-7}
            \multicolumn{1}{c}{\scriptsize Train $\setminus$ Test} & \multicolumn{1}{c}{T1} & \multicolumn{1}{c}{T2} & \multicolumn{1}{c}{T3} & \multicolumn{1}{c}{T4} & \multicolumn{1}{c}{T5} & \multicolumn{1}{c}{T6} \\ \hline
            T1 & 44.4 & \multicolumn{5}{c}{} \\ \cline{2-2} \cline{3-3}
            T2 & 43.2 & 64.5 & \multicolumn{4}{c}{} \\ \cline{2-3} \cline{4-4}
            T3 & 18.4 & 52.2 & 21.3 & \multicolumn{3}{c}{} \\ \cline{2-4} \cline{5-5}
            T4 & 35.4 & 51.0 & 19.2 & 54.2 & \multicolumn{2}{c}{} \\ \cline{2-5} \cline{6-6}
            T5 & 35.8 & 43.8 & 14.7 & 44.8 & 29.6 & \multicolumn{1}{c}{} \\ \cline{2-6} \cline{7-7}
            T6 & 33.6 & 8.1 & 14.4 & 44.2 & 22.2 & 41.4 \\ \cline{2-7}
        \end{tabular}
        \quad 
         \begin{tabular}{r |c|c|c|c|c|c|}
    \multicolumn{1}{c}{} & \multicolumn{6}{c}{\textbf{SETA (ours)}} \\ \cline{1-7}
    \multicolumn{1}{c}{\scriptsize Train $\setminus$ Test} & \multicolumn{1}{c}{T1} & \multicolumn{1}{c}{T2} & \multicolumn{1}{c}{T3} & \multicolumn{1}{c}{T4} & \multicolumn{1}{c}{T5} & \multicolumn{1}{c}{T6} \\ \hline
    T1 & 47.69 & \multicolumn{5}{c}{} \\ \cline{2-2} \cline{3-3}
    T2 & 43.80 & 65.93 & \multicolumn{4}{c}{} \\ \cline{2-3} \cline{4-4}
    T3 & 33.60 & 49.39 & 43.83 & \multicolumn{3}{c}{} \\ \cline{2-4} \cline{5-5}
    T4 & 31.60 & 31.45 & 24.68& 34.34 & \multicolumn{2}{c}{} \\ \cline{2-5} \cline{6-6}
    T5 & 32.60 & 49.19 & 23.91 & 48.18 & 28.57 & \multicolumn{1}{c}{} \\ \cline{2-6} \cline{7-7}
    T6 & 27.31 & 26.41 & 20.81 & 41.63 & 22.55 & 33.62 \\ \cline{2-7}
\end{tabular}
    }
\end{table*}
\paragraph{Analysis of Stability and Plasticity:}
Table~\ref{tbl:task_wise_performance_qwen3_4b} tracks sequential
performance on Qwen3-4B, where diagonal entries reflect initial
plasticity and off-diagonal decay reflects forgetting. SETA exhibits
a stable retention profile: Task~1 accuracy rises to 63.64\% by the
final step, while EWC and I-LoRA collapse substantially, confirming
that the Split-on-Share mechanism actively preserves early task
knowledge. On plasticity, SETA achieves 75.90\% on Task~4 and
sustains high retention on earlier tasks throughout the sequence.
Table~\ref{tbl:task_wise_performance} confirms that this
stability--plasticity balance generalises to LLaMA-2 7B.
\subsection{Layer-wise Selection Across Architectures}
\label{app:phase0:layer}
\begin{figure}[htbp]
  \centering
  \includegraphics[width=\linewidth]{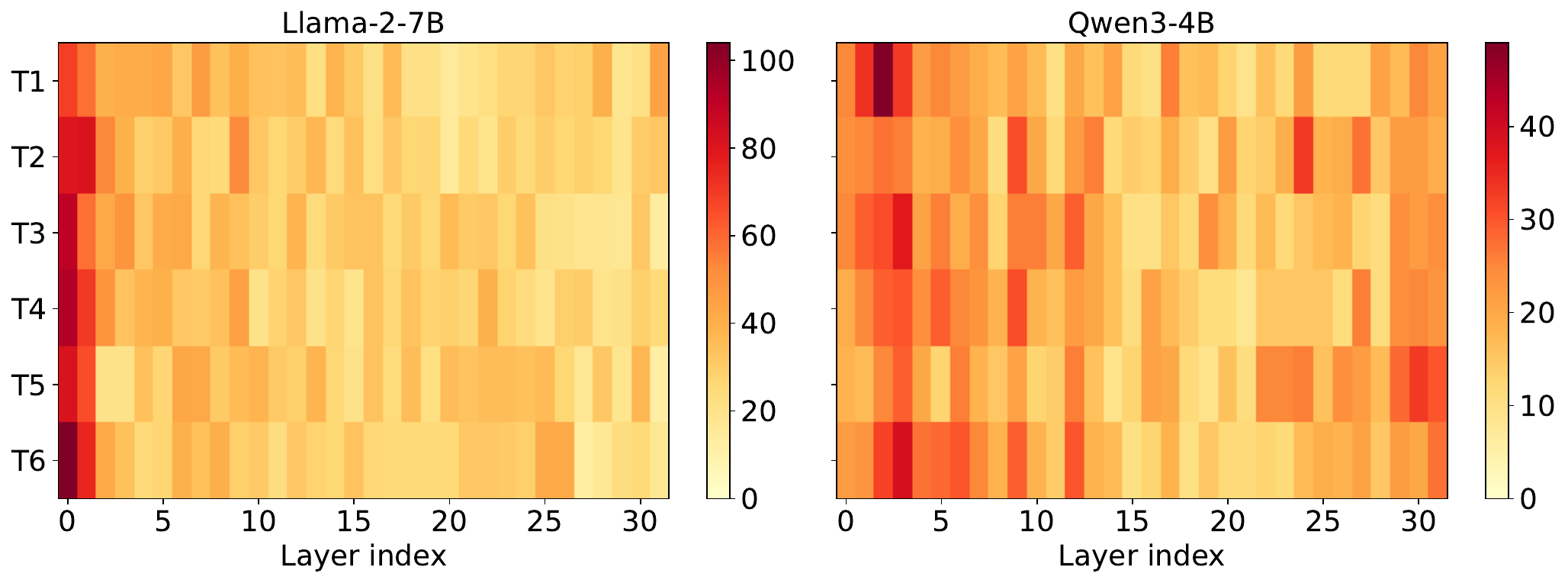}
  \caption{Unique block count per transformer layer and task for LLaMA-2 7B and Qwen3-4B. Darker cells indicate more unique blocks at that layer. Both architectures show non-uniform, task-sensitive selection patterns without architecture-specific tuning.}
  \label{fig:layer_coverage}
\end{figure}
\begin{figure}[htbp]
    \centering
    \includegraphics[width=\linewidth]{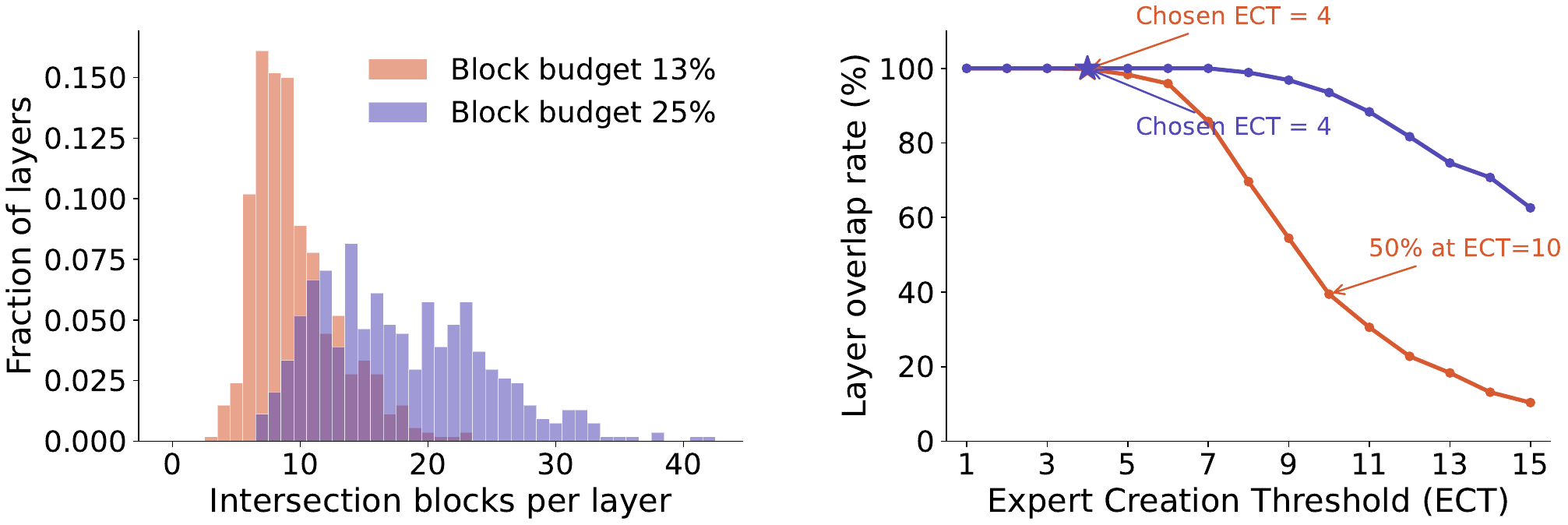}
    \caption{Effect of block budget fraction on expert creation for
  Qwen3-4B. Left: intersection block count distributions for 13\% and
  25\% budgets. Right: layer overlap rate as a function of ECT; stars
  mark the chosen ECT and annotations mark the 50\% approval knee.}
    \label{fig:budget_analysis}
\end{figure}
Figure~\ref{fig:budget_analysis} reveals that increasing the block budget 
from 13\% to 25\% shifts the intersection distribution rightward, 
producing larger per-layer overlaps between tasks. As a consequence, the 
same ECT becomes more permissive under a larger budget: the 13\% budget 
curve reaches 50\% approval at ECT$=10$, while the 25\% budget curve 
remains above 60\% even at ECT$=15$. This confirms that ECT must be 
calibrated relative to the block budget: a fixed absolute threshold 
approves far more splits under a larger budget, potentially creating 
spurious shared experts from incidental overlap. The chosen ECT values of $4$ for the 13\% budget and $8$ for the 25\% budget each represent a principled operating point at the elbow of the approval curve, rejecting noise while preserving genuine knowledge sharing between tasks.
\subsection{Unique Sub-space Discovery Rate}
\label{app:phase0:rate}
\begin{figure}[htbp]
  \centering
  \includegraphics[width=\linewidth]{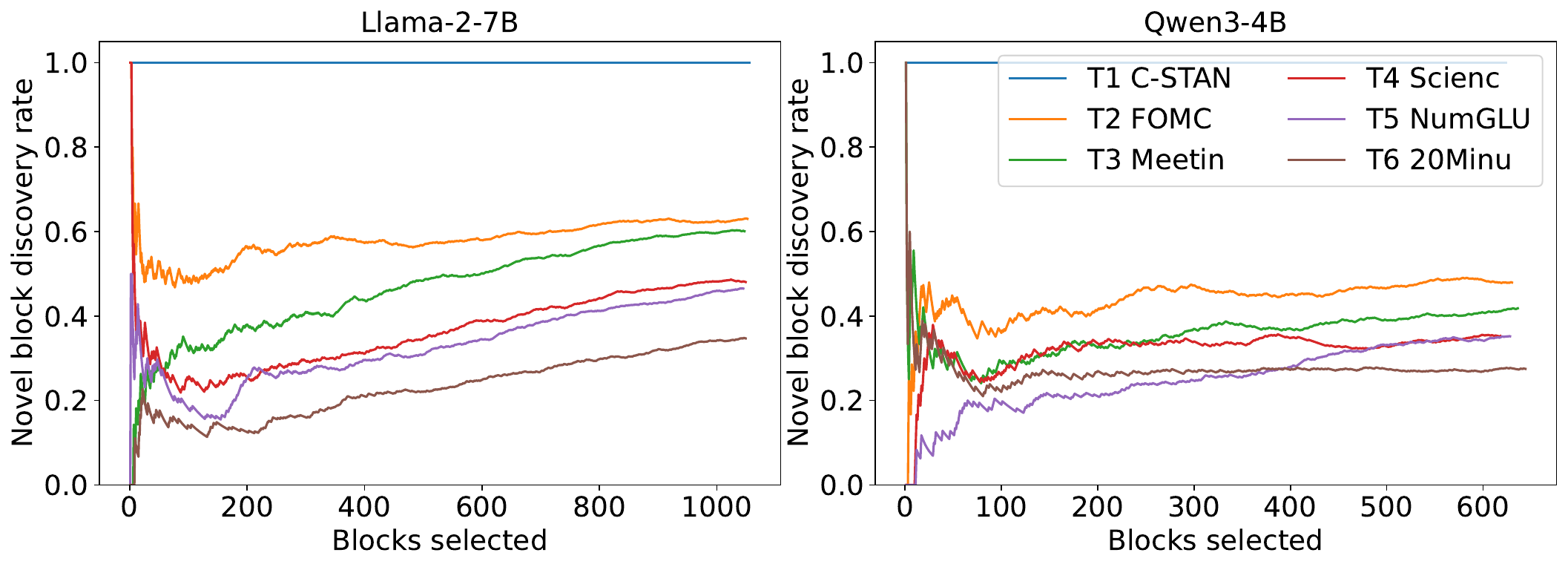}
  \caption{Novel block discovery rate during FFT selection for LLaMA-2 7B and Qwen3-4B. T1 maintains a rate of 1.0 throughout; later tasks decay progressively faster as prior coverage accumulates.}
  \label{fig:novel_rate}
\end{figure}
Figure~\ref{fig:novel_rate} shows that as the task sequence grows,
each new task finds progressively less novel ground to cover.
The monotonic degradation of later-task discovery rates is consistent
across both LLaMA-2 7B and Qwen3-4B, confirming sub-linear capacity
expansion as a model-agnostic property of the SETA selection mechanism.
\subsubsection{Parameter Efficiency}
In Table~\ref{tab:trainable_parameter}, our method maintains a minimal memory footprint. The trainable parameters (\#TP) increase marginally from 6.34M (Task 1) to 8.10M (Task 6), representing just 0.98\% to 1.25\% of the total model capacity. Notably, this remains comparable to or lower than I-LoRA, which requires approximately 8.39M trainable parameters with rank 8. This confirms effective adaptation while modifying $\approx$1\% of the parameters. However, this extreme parameter sparsity introduces inherent network saturation limits. As the model adapts to sequential tasks, the strictly bounded capacity forces a heavy reliance on the frozen shared core, eventually creating a plasticity bottleneck that constrains forward transfer for future tasks.
\begin{table}[htbp]
\centering
\caption{Parameter efficiency analysis showing the number of trainable parameters (\#TP) in millions and their percentage relative to LLaMA-2 7B; I-LoRA (rank 8) requires approximately 8.39M parameters for reference.}
\label{tab:trainable_parameter}
\setlength{\tabcolsep}{3.8pt} 
\renewcommand{\arraystretch}{1.1} 
\small
\begin{tabular}{lcccccc}
\toprule
\textbf{Metrics} & \textbf{T1} & \textbf{T2} & \textbf{T3} & \textbf{T4} & \textbf{T5} & \textbf{T6}  \\
\midrule
\#TP (M) $\downarrow$ & 6.34 & 6.33 & 6.62 & 6.53 & 7.16 & 8.10  \\
TP (\%) $\downarrow$  & 0.98 & 0.98 & 1.03 & 1.01  & 1.11 & 1.25 \\
\bottomrule
\end{tabular}
\end{table}
\section{Related work: PEFT in CL}
Fine-tuning large language models (LLMs) is computationally demanding due to their massive parameter counts. Parameter-efficient fine-tuning (PEFT) methods address this challenge by updating only a small subset of parameters while keeping pre-trained weights frozen~\cite{li2021prefix}. A variety of PEFT techniques have been proposed, including Adapter Tuning~\cite{houlsby2019parameter}, BitFit~\cite{zaken2022bitfit}, Prefix Tuning~\cite{li2021prefix}, Prompt Tuning~\cite{lester2021power}, and Low-Rank Adaptation (LoRA)~\cite{hu2022lora}. Extensions such as ReLoRA~\cite{lialin2024relora} and RankAdapter~\cite{zhou2024rankadaptor} improve memory efficiency and dynamically adjust ranks, though they lack formal guarantees. AdaZeta~\cite{yang2024adazeta} introduces zeroth-order optimization with convergence guarantees, while other methods~\cite{gao2024adaptive} explore adaptive-rank strategies without theoretical proofs. LoRA has also been combined with Mixture-of-Experts architectures~\cite{li2024mixlora, huang2024mixture}, as in AdaMoLE~\cite{liu2024adamole}, enabling dynamic expert selection, and with Neural Architecture Search for LLM compression~\cite{munoz2025low}. Another line of work leverages sparsity, such as Sparse Matrix Tuning (SMT)~\cite{he2025smt}, which selects task-relevant sub-matrices.
\section{Conclusion}
We presented SETA, a framework for mitigating catastrophic forgetting in large language models under continual learning. By separating task-specific and shared knowledge within a dynamically evolving Mixture-of-Experts architecture, SETA isolates conflicting updates while preserving reusable representations through elastic weight anchoring. Experimental results show that this structural decomposition enables task-agnostic inference and achieves competitive or superior performance relative to state-of-the-art continual learning baselines across diverse reasoning benchmarks. 
% Future work includes scaling SETA to longer sequences and larger models while exploring adaptive growth. 
\section*{Acknowledgments}
This research was supported by the Argonne Given Internship Program and U.S. Department of Energy, Office of Science, Advanced Scientific Computing Research, under contract number DE-AC02-06CH11357. This research used resources of the Argonne Leadership Computing Facility at Argonne National Laboratory, which is supported by the Office of Science of the U.S. Department of Energy, Office of Science, under contract number DE-AC02-06CH11357. We appreciate the support from the ISU Dean's Emerging Faculty Leaders Award. We also thank the National Center for Supercomputing Applications for providing Delta GPUs through allocation CIS240604 from the Advanced Cyberinfrastructure Coordination Ecosystem: Services \& Support (ACCESS) program.
\section*{Limitations}
While SETA shows strong continual learning performance, several limitations remain. 
Our experiments cover models up to 14B parameters, so how expert specialization 
and routing scale to larger models is still unknown. SETA has also been validated 
only on standard attention-based transformers; its behavior on alternative 
architectures remains an open question.
Our evaluation follows a fixed task ordering across six benchmarks. Longer 
sequences, greater task diversity, and different orderings may reveal new 
behaviors or failure modes. Finally, as tasks grow, the shared expert pool 
expands without any pruning or consolidation mechanism, which may limit 
scalability in long-horizon settings. We leave these directions as important 
avenues for future work.

\bibliography{custom}

@article{brown2020language,
  title={Language models are few-shot learners},
  author={Brown, Tom and Mann, Benjamin and Ryder, Nick and Subbiah, Melanie and Kaplan, Jared D and Dhariwal, Prafulla and Neelakantan, Arvind and Shyam, Pranav and Sastry, Girish and Askell, Amanda and others},
  journal={Advances in neural information processing systems},
  volume={33},
  pages={1877--1901},
  year={2020}
}

@article{raffel2020exploring,
  title={Exploring the limits of transfer learning with a unified text-to-text transformer},
  author={Raffel, Colin and Shazeer, Noam and Roberts, Adam and Lee, Katherine and Narang, Sharan and Matena, Michael and Zhou, Yanqi and Li, Wei and Liu, Peter J},
  journal={Journal of machine learning research},
  volume={21},
  number={140},
  pages={1--67},
  year={2020}
}

@article{parisi2019continual,
  title={Continual lifelong learning with neural networks: A review},
  author={Parisi, German I and Kemker, Ronald and Part, Jose L and Kanan, Christopher and Wermter, Stefan},
  journal={Neural networks},
  volume={113},
  pages={54--71},
  year={2019},
  publisher={Elsevier}
}

@article{li2017learning,
  title={Learning without forgetting},
  author={Li, Zhizhong and Hoiem, Derek},
  journal={IEEE transactions on pattern analysis and machine intelligence},
  volume={40},
  number={12},
  pages={2935--2947},
  year={2017},
  publisher={IEEE}
}

@article{chaudhry2019tiny,
  title={On tiny episodic memories in continual learning},
  author={Chaudhry, Arslan and Rohrbach, Marcus and Elhoseiny, Mohamed and Ajanthan, Thalaiyasingam and Dokania, Puneet K and Torr, Philip HS and Ranzato, Marc'Aurelio},
  journal={arXiv preprint arXiv:1902.10486},
  year={2019}
}

@article{kirkpatrick2017overcoming,
  title={Overcoming catastrophic forgetting in neural networks},
  author={Kirkpatrick, James and Pascanu, Razvan and Rabinowitz, Neil and Veness, Joel and Desjardins, Guillaume and Rusu, Andrei A and Milan, Kieran and Quan, John and Ramalho, Tiago and Grabska-Barwinska, Agnieszka and others},
  journal={Proceedings of the national academy of sciences},
  volume={114},
  number={13},
  pages={3521--3526},
  year={2017},
  publisher={National Academy of Sciences}
}

@article{rusu2016progressive,
  title={Progressive neural networks},
  author={Rusu, Andrei A and Rabinowitz, Neil C and Desjardins, Guillaume and Soyer, Hubert and Kirkpatrick, James and Kavukcuoglu, Koray and Pascanu, Razvan and Hadsell, Raia},
  journal={arXiv preprint arXiv:1606.04671},
  year={2016}
}

@article{han2015learning,
  title={Learning both weights and connections for efficient neural network},
  author={Han, Song and Pool, Jeff and Tran, John and Dally, William},
  journal={Advances in neural information processing systems},
  volume={28},
  year={2015}
}

@inproceedings{evci2020rigging,
  title={Rigging the lottery: Making all tickets winners},
  author={Evci, Utku and Gale, Trevor and Menick, Jacob and Castro, Pablo Samuel and Elsen, Erich},
  booktitle={International conference on machine learning},
  pages={2943--2952},
  year={2020},
  organization={PMLR}
}

@inproceedings{li2025analyzing,
  title={Analyzing and reducing catastrophic forgetting in parameter efficient tuning},
  author={Li, Xinlong and Ren, Weijieying and Qin, Wei and Wang, Lei and Zhao, Tianxiang and Hong, Richang},
  booktitle={ICASSP 2025-2025 IEEE International Conference on Acoustics, Speech and Signal Processing (ICASSP)},
  pages={1--5},
  year={2025},
  organization={IEEE}
}

@article{wu2024continual,
  title={Continual learning for large language models: A survey},
  author={Wu, Tongtong and Luo, Linhao and Li, Yuan-Fang and Pan, Shirui and Vu, Thuy-Trang and Haffari, Gholamreza},
  journal={arXiv preprint arXiv:2402.01364},
  year={2024}
}

@article{hu2022lora,
  title={Lora: Low-rank adaptation of large language models.},
  author={Hu, Edward J and Shen, Yelong and Wallis, Phillip and Allen-Zhu, Zeyuan and Li, Yuanzhi and Wang, Shean and Wang, Liang and Chen, Weizhu and others},
  journal={Iclr},
  volume={1},
  number={2},
  pages={3},
  year={2022}
}

@article{luo2025empirical,
  title={An empirical study of catastrophic forgetting in large language models during continual fine-tuning},
  author={Luo, Yun and Yang, Zhen and Meng, Fandong and Li, Yafu and Zhou, Jie and Zhang, Yue},
  journal={IEEE Transactions on Audio, Speech and Language Processing},
  year={2025},
  publisher={IEEE}
}

@inproceedings{ren2024analyzing,
  title={Analyzing and reducing catastrophic forgetting in parameter efficient tuning},
  author={Li, Xinlong and Ren, Weijieying and Qin, Wei and Wang, Lei and Zhao, Tianxiang and Hong, Richang},
  booktitle={ICASSP 2025-2025 IEEE International Conference on Acoustics, Speech and Signal Processing (ICASSP)},
  pages={1--5},
  year={2025},
  organization={IEEE}
}

@inproceedings{lopez2017gradient,
  title={Gradient episodic memory for continual learning},
  author={Lopez-Paz, David and Ranzato, Marc'Aurelio},
  booktitle={Advances in Neural Information Processing Systems},
  volume={30},
  year={2017}
}

@inproceedings{farajtabar2020orthogonal,
  title={Orthogonal Gradient Descent for Continual Learning},
  author={Farajtabar, Mehrdad and Azizan, Navid and Mott, Alex and Li, Ang},
  booktitle={International Conference on Artificial Intelligence and Statistics},
  pages={3762--3773},
  year={2020},
  organization={PMLR}
}

@article{razdaibiedina2023progressive,
  title={Progressive prompts: Continual learning for language models},
  author={Razdaibiedina, Anastasia and Mao, Yuning and Hou, Rui and Khabsa, Madian and Lewis, Mike and Almahairi, Amjad},
  journal={arXiv preprint arXiv:2301.12314},
  year={2023}
}

@inproceedings{li2021prefix,
  title={Prefix-tuning: Optimizing continuous prompts for generation},
  author={Li, Xiang Lisa and Liang, Percy},
  booktitle={Proceedings of the 59th Annual Meeting of the Association for Computational Linguistics and the 11th International Joint Conference on Natural Language Processing (Volume 1: Long Papers)},
  pages={4582--4597},
  year={2021}
}

@incollection{mccloskey1989catastrophic,
  title={Catastrophic interference in connectionist networks: The sequential learning problem},
  author={McCloskey, Michael and Cohen, Neal J},
  booktitle={Psychology of learning and motivation},
  volume={24},
  pages={109--165},
  year={1989},
  publisher={Elsevier}
}

@inproceedings{rebuffi2017icarl,
  title={icarl: Incremental classifier and representation learning},
  author={Rebuffi, Sylvestre-Alvise and Kolesnikov, Alexander and Sperl, Georg and Lampert, Christoph H},
  booktitle={Proceedings of the IEEE conference on Computer Vision and Pattern Recognition},
  pages={2001--2010},
  year={2017}
}

@inproceedings{li2024ccpeft,
  title={Customizable Combination of Parameter-Efficient Modules for Multi-Task Learning},
  author={Li, Han and Xu, Yuchen and others},
  booktitle={Proceedings of EMNLP},
  year={2024}
}

@inproceedings{han2025slim,
  title={Slim: Let llm learn more and forget less with soft lora and identity mixture},
  author={Han, Jiayi and Du, Liang and Du, Hongwei and Zhou, Xiangguo and Wu, Yiwen and Zhang, Yuanfang and Zheng, Weibo and Han, Donghong},
  booktitle={Proceedings of the 2025 Conference of the Nations of the Americas Chapter of the Association for Computational Linguistics: Human Language Technologies (Volume 1: Long Papers)},
  pages={4792--4804},
  year={2025}
}

@inproceedings{houlsby2019parameter,
  title={Parameter-efficient transfer learning for NLP},
  author={Houlsby, Neil and Giurgiu, Andrei and Jastrzebski, Stanislaw and Morrone, Bruna and De Laroussilhe, Quentin and Gesmundo, Andrea and Attariyan, Mona and Gelly, Sylvain},
  booktitle={International conference on machine learning},
  pages={2790--2799},
  year={2019},
  organization={PMLR}
}

@inproceedings{zaken2022bitfit,
  title={Bitfit: Simple parameter-efficient fine-tuning for transformer-based masked language-models},
  author={Zaken, Elad Ben and Goldberg, Yoav and Ravfogel, Shauli},
  booktitle={Proceedings of the 60th Annual Meeting of the Association for Computational Linguistics (Volume 2: Short Papers)},
  pages={1--9},
  year={2022}
}

@inproceedings{lester2021power,
  title={The power of scale for parameter-efficient prompt tuning},
  author={Lester, Brian and Al-Rfou, Rami and Constant, Noah},
  booktitle={Proceedings of the 2021 conference on empirical methods in natural language processing},
  pages={3045--3059},
  year={2021}
}

@inproceedings{lialin2024relora,
  title={Relora: High-rank training through low-rank updates},
  author={Lialin, Vladislav and Muckatira, Sherin and Shivagunde, Namrata and Rumshisky, Anna},
  booktitle={International Conference on Learning Representations},
  volume={2024},
  pages={49405--49421},
  year={2024}
}

@article{zhou2024rankadaptor,
  title={Rankadaptor: Hierarchical dynamic low-rank adaptation for structural pruned llms},
  author={Zhou, Changhai and Han, Shijie and Zhang, Shiyang and Weng, Shichao and Liu, Zekai and Jin, Cheng},
  journal={arXiv preprint arXiv:2406.15734},
  year={2024},
  publisher={Jun}
}

@inproceedings{yang2024adazeta,
  title={Adazeta: Adaptive zeroth-order tensor-train adaption for memory-efficient large language models fine-tuning},
  author={Yang, Yifan and Zhen, Kai and Banijamali, Ershad and Mouchtaris, Athanasios and Zhang, Zheng},
  booktitle={Proceedings of the 2024 Conference on Empirical Methods in Natural Language Processing},
  pages={977--995},
  year={2024}
}

@inproceedings{gao2024adaptive,
  title={Adaptive rank selections for low-rank approximation of language models},
  author={Gao, Shangqian and Hua, Ting and Hsu, Yen-Chang and Shen, Yilin and Jin, Hongxia},
  booktitle={Proceedings of the 2024 Conference of the North American Chapter of the Association for Computational Linguistics: Human Language Technologies (Volume 1: Long Papers)},
  pages={227--241},
  year={2024}
}

@article{li2024mixlora,
  title={Mixlora: Enhancing large language models fine-tuning with lora-based mixture of experts},
  author={Li, Dengchun and Ma, Yingzi and Wang, Naizheng and Ye, Zhengmao and Cheng, Zhiyuan and Tang, Yinghao and Zhang, Yan and Duan, Lei and Zuo, Jie and Yang, Cal and others},
  journal={arXiv preprint arXiv:2404.15159},
  year={2024}
}

@inproceedings{huang2024mixture,
  title={Mixture of lora experts},
  author={Huang, Shaohan and Wei, Furu and others},
  booktitle={International Conference on Learning Representations},
  volume={2024},
  pages={47302--47318},
  year={2024}
}

@article{liu2024adamole,
  title={Adamole: Fine-tuning large language models with adaptive mixture of low-rank adaptation experts},
  author={Liu, Zefang and Luo, Jiahua},
  journal={arXiv preprint arXiv:2405.00361},
  year={2024}
}

@article{munoz2025low,
  title={Low-rank adapters meet neural architecture search for llm compression},
  author={Mu{\~n}oz, J Pablo and Yuan, Jinjie and Jain, Nilesh},
  journal={arXiv preprint arXiv:2501.16372},
  year={2025}
}

@inproceedings{he2025smt,
  title={Smt: Fine-tuning large language models with sparse matrices},
  author={He, Haoze and Li, Juncheng and Jiang, Xuan and Miller, Heather},
  booktitle={International Conference on Learning Representations},
  volume={2025},
  pages={32369--32393},
  year={2025}
}

@article{shazeer2017outrageously,
  title={Outrageously large neural networks: The sparsely-gated mixture-of-experts layer},
  author={Shazeer, Noam and Mirhoseini, Azalia and Maziarz, Krzysztof and Davis, Andy and Le, Quoc and Hinton, Geoffrey and Dean, Jeff},
  journal={arXiv preprint arXiv:1701.06538},
  year={2017}
}

@article{lepikhin2020gshard,
  title={Gshard: Scaling giant models with conditional computation and automatic sharding},
  author={Lepikhin, Dmitry and Lee, HyoukJoong and Xu, Yuanzhong and Chen, Dehao and Firat, Orhan and Huang, Yanping and Krikun, Maxim and Shazeer, Noam and Chen, Zhifeng},
  journal={arXiv preprint arXiv:2006.16668},
  year={2020}
}

@article{fedus2022switch,
  title={Switch transformers: Scaling to trillion parameter models with simple and efficient sparsity},
  author={Fedus, William and Zoph, Barret and Shazeer, Noam},
  journal={Journal of Machine Learning Research},
  volume={23},
  number={120},
  pages={1--39},
  year={2022}
}

@inproceedings{lewis2021base,
  title={Base layers: Simplifying training of large, sparse models},
  author={Lewis, Mike and Bhosale, Shruti and Dettmers, Tim and Goyal, Naman and Zettlemoyer, Luke},
  booktitle={International Conference on Machine Learning},
  pages={6265--6274},
  year={2021},
  organization={PMLR}
}

@article{roller2021hash,
  title={Hash layers for large sparse models},
  author={Roller, Stephen and Sukhbaatar, Sainbayar and Weston, Jason and others},
  journal={advances in neural information processing systems},
  volume={34},
  pages={17555--17566},
  year={2021}
}

@article{zhou2022mixture,
  title={Mixture-of-experts with expert choice routing},
  author={Zhou, Yanqi and Lei, Tao and Liu, Hanxiao and Du, Nan and Huang, Yanping and Zhao, Vincent and Dai, Andrew M and Le, Quoc V and Laudon, James and others},
  journal={Advances in Neural Information Processing Systems},
  volume={35},
  pages={7103--7114},
  year={2022}
}

@article{lu2022learn,
  title={Learn to explain: Multimodal reasoning via thought chains for science question answering},
  author={Lu, Pan and Mishra, Swaroop and Xia, Tanglin and Qiu, Liang and Chang, Kai-Wei and Zhu, Song-Chun and Tafjord, Oyvind and Clark, Peter and Kalyan, Ashwin},
  journal={Advances in neural information processing systems},
  volume={35},
  pages={2507--2521},
  year={2022}
}

@inproceedings{shah2023trillion,
  title={Trillion dollar words: A new financial dataset, task \& market analysis},
  author={Shah, Agam and Paturi, Suvan and Chava, Sudheer},
  booktitle={Proceedings of the 61st Annual Meeting of the Association for Computational Linguistics (Volume 1: Long Papers)},
  pages={6664--6679},
  year={2023}
}

@inproceedings{hu2023meetingbank,
  title={Meetingbank: A benchmark dataset for meeting summarization},
  author={Hu, Yebowen and Ganter, Timothy and Deilamsalehy, Hanieh and Dernoncourt, Franck and Foroosh, Hassan and Liu, Fei},
  booktitle={Proceedings of the 61st Annual Meeting of the Association for Computational Linguistics (Volume 1: Long Papers)},
  pages={16409--16423},
  year={2023}
}

@inproceedings{zhao2023c,
  title={C-STANCE: A large dataset for Chinese zero-shot stance detection},
  author={Zhao, Chenye and Li, Yingjie and Caragea, Cornelia},
  booktitle={Proceedings of the 61st Annual Meeting of the Association for Computational Linguistics (Volume 1: Long Papers)},
  pages={13369--13385},
  year={2023}
}

@inproceedings{kew202320,
  title={20 Minuten: A Multi-task News Summarisation Dataset for German},
  author={Kew, Tannon and Kostrzewa, Marek and Ebling, Sarah},
  booktitle={Proceedings of the 8th edition of the Swiss Text Analytics Conference},
  pages={1--13},
  year={2023}
}

@inproceedings{mishra2022numglue,
  title={NumGLUE: A suite of fundamental yet challenging mathematical reasoning tasks},
  author={Mishra, Swaroop and Mitra, Arindam and Varshney, Neeraj and Sachdeva, Bhavdeep and Clark, Peter and Baral, Chitta and Kalyan, Ashwin},
  booktitle={Proceedings of the 60th Annual Meeting of the Association for Computational Linguistics (Volume 1: Long Papers)},
  pages={3505--3523},
  year={2022}
}

\appendix
\section*{Appendix}
\section{Related Works}
\label{appendix:related_works}
% \todo[inline]{This section could be moved to the Appendix and mentioned in the outline.}
\subsection{Continual Learning}
Continual learning (CL) seeks to train models on sequential tasks without catastrophic forgetting ~\citep{mccloskey1989catastrophic}, where newly acquired knowledge overwrites previously learned representations. 
Classical CL methods include regularization-based approaches such as EWC~\citep{kirkpatrick2017overcoming} and LwF~\citep{li2017learning}, rehearsal-based replay~\citep{rebuffi2017icarl}, and architectural expansion~\citep{rusu2016progressive}, though these often require large memory buffers or suffer from scalability issues. Recent advances extend CL to Transformer-based and large language models (LLMs) using parameter-efficient tuning (PEFT) modules. 
For instance, I-LoRA~\citep{li2025analyzing} mitigates forgetting by maintaining dual LoRA adapters, a fast (short-term) and a slow (long-term) memory, interpolated to balance plasticity and stability. 
SLiM~\citep{han2025slim} introduces a soft mixture of LoRA and identity mappings, dynamically weighting learned and frozen components to enhance retention while preserving efficiency. 
Meanwhile, CC-PEFT~\citep{li2024ccpeft} generalizes modular parameter-efficient tuning by combining multiple PEFT adapters in a customizable manner for multi-task and continual scenarios. 

\subsection{Mixture-of-Experts (MoE)} 
Traditional dense models process every input token with all parameters, making computation scale linearly with model size. 
Mixture-of-Experts (MoE) improves efficiency by activating only a subset of parameters, first shown effective in LSTMs \citep{shazeer2017outrageously} and later integrated into Transformers \citep{lepikhin2020gshard, fedus2022switch}. 
Subsequent work addressed load imbalance with routing strategies \citep{lewis2021base, roller2021hash, zhou2022mixture}, and recent decoder-only variants such as UL2-MoE and Mixtral further advanced scalability. 
\section{Ablation: Routing-Aware Gating Regularisation}
\label{app:ablation:gating}
Table~\ref{tbl:qwen34b_gating_ablation} report the sequential task-accuracy matrices for Qwen3-4B under identical
settings except for the routing-aware gating loss $\mathcal{L}_{\text{gate}}$.
T1 and T2 rows are identical across both configurations because shared experts
only exist from T3 onward; the gating loss therefore has no effect until the
first Split-on-Share event. Enabling $\mathcal{L}_{\text{gate}}$ improves average accuracy at T6 from
38.48 to 43.30  and reduces forgetting as measured by BWT
from $-15.70$ to $-12.67$. The gains are concentrated in two areas.
\begin{table*}[htbp]
\centering
\caption{Qwen3-4B sequential accuracy matrix under dynamic $\lambda$,
  with routing-aware gating disabled and enabled. Diagonal entries
  (bold) measure plasticity; off-diagonal entries measure retention.}
\label{tbl:qwen34b_gating_ablation}
\small
\begin{tabular}{lcccccccc}
\toprule
& \multicolumn{7}{c}{\textbf{Test Task}} \\
\cmidrule(lr){2-8}
\textbf{Train} & \textbf{T1} & \textbf{T2} & \textbf{T3} & \textbf{T4} & \textbf{T5} & \textbf{T6} & \textbf{OP $\uparrow$} & \textbf{BWT$\uparrow$} \\
\midrule
\multicolumn{9}{l}{\textit{Routing-Aware Gating: Disabled}} \\
\midrule
T1 & 57.01 & & & & & & & \\
T2 & 57.99 & 67.54 & & & & & & \\
T3 & 50.61 & 56.25 & 38.04 & & & & & \\
T4 & 37.92 & 28.83 & 23.56 & 79.29 & & & & \\
T5 & 49.79 & 28.83 & 20.95 & 73.44 & 34.62 & & & \\
T6 & 53.64 & 32.06 & 19.04 & 78.00 & 15.28 & 32.84 & 38.48 & $-15.70$ \\
\midrule
\multicolumn{9}{l}{\textit{Routing-Aware Gating: Enabled (SETA - ours)}} \\
\midrule
T1 & 57.01 & & & & & & & \\
T2 & 57.99 & 67.54 & & & & & & \\
T3 & 52.12 & 56.45 & 46.69 & & & & & \\
T4 & 37.29 & 26.00 & 24.16 & 75.90 & & & & \\
T5 & 59.00 & 38.91 & 22.83 & 83.01 & 41.03 & & & \\
T6 & 63.64 & 33.00 & 21.42 & 80.77 & 25.97 & 34.98 & \textbf{43.30} & $\mathbf{-12.67}$ \\
\bottomrule
\end{tabular}
\end{table*}
\paragraph{Retention of early tasks:} C-STANCE accuracy at the final step rises
from 53.64 to 63.64, a gain of 10.0 points, while NumGLUE-cm retention at T6
improves from 15.28 to 25.97. These are precisely the tasks
whose shared experts accumulate the highest $\omega_i$ weights and therefore
receive the strongest gate-steering signal, confirming that
$\mathcal{L}_{\text{gate}}$ acts where it is most needed.

\textbf{Plasticity on intermediate tasks:} MeetingBank diagonal accuracy
improves from 38.04 to 46.69 (+8.65 points) and NumGLUE-cm from 34.62 to
41.03 (+6.41 points). Because the gating loss reduces routing probability
toward saturated shared experts, the gating network more readily directs
new-task inputs toward available unique experts, preserving plasticity. The
modest diagonal drop on ScienceQA is consistent with
the same mechanism: stricter routing reduces the task's access to broadly
shared representations, a known stability--plasticity trade-off that the
overall BWT improvement confirms is net beneficial.

Together, these results show that routing-aware gating and weight
regularisation are complementary: $\mathcal{L}_{\text{reg}}$ corrects drift
after gradient updates have occurred, while $\mathcal{L}_{\text{gate}}$
prevents the gating network from routing new-task inputs into at-risk experts
in the first place, providing a two-layer defence at negligible additional
cost since both losses share the same $\lambda_i(t)$ coefficients and are
evaluated in a single forward pass.

\section{Empirical Analysis and Design Choices}

This section presents the structural analysis of SETA's gradient-guided
block selection across two architectures, LLaMa-2 7B, LLaMA-3 8B and Qwen3-4B,
using the TRACE benchmark tasks. 
\subsection{Attention Sub-block Scaling Analysis}

\begin{table*}[htbp]
\centering
\caption{Evolution of sub-block uniqueness statistics for Qwen3-4B under
two block budget fractions and their respective Expert Creation Thresholds
(ECT). Total Blocks = cumulative raw blocks selected by FFT. Total Shared
= blocks overlapping across tasks passing ECT.}
\label{tab:uniqueness_qwen3}
\begin{tabular}{clrrrrrrrr}
\toprule
\textbf{Seq} & \textbf{Task} & \textbf{Total Blocks} & \textbf{Total Shared}
& \textbf{T1} & \textbf{T2} & \textbf{T3} & \textbf{T4} & \textbf{T5} & \textbf{T6} \\
\midrule
\multicolumn{10}{l}{\textit{Block budget 13\% (ECT = 4)}} \\
\midrule
1 & C-STANCE    & 720  & 0   & 720 & 0   & 0   & 0   & 0   & 0   \\
2 & FOMC        & 1449 & 388 & 332 & 341 & 0   & 0   & 0   & 0   \\
3 & MeetingBank & 2180 & 530 & 251 & 280 & 315 & 0   & 0   & 0   \\
4 & ScienceQA   & 2908 & 672 & 207 & 239 & 258 & 250 & 0   & 0   \\
5 & NumGLUE-cm  & 3640 & 796 & 180 & 210 & 223 & 217 & 234 & 0   \\
6 & 20Minuten   & 4367 & 922 & 159 & 185 & 194 & 182 & 218 & 195 \\
\midrule
\multicolumn{10}{l}{\textit{Block budget 25\% (ECT = 8)}} \\
\midrule
1 & C-STANCE    & 1440 & 0    & 1440 & 0   & 0   & 0   & 0   & 0   \\
2 & FOMC        & 2862 & 744  & 696  & 678 & 0   & 0   & 0   & 0   \\
3 & MeetingBank & 4282 & 1140 & 489  & 489 & 598 & 0   & 0   & 0   \\
4 & ScienceQA   & 5702 & 1527 & 364  & 378 & 447 & 431 & 0   & 0   \\
5 & NumGLUE-cm  & 7124 & 1906 & 281  & 277 & 352 & 331 & 320 & 0   \\
6 & 20Minuten   & 8545 & 2225 & 225  & 224 & 269 & 258 & 266 & 256 \\
\bottomrule
\end{tabular}
\end{table*}

Table~\ref{tab:uniqueness_qwen3} presents uniqueness statistics for
Qwen3-4B under two block budget configurations.
Under the 13\% budget, the shared pool grows from 0 to 922 blocks across
six tasks, with T1 retaining only 22\% of its original unique blocks by T6.
Under the 25\% budget, shared blocks reach 26\% of total by T6, with T1
retaining 16\%. The larger budget thus drives proportionally stronger sharing
despite the stricter ECT, confirming that ECT must be calibrated relative to
the block budget rather than set as a fixed constant.

\subsection{Block Selection Budget and Efficiency}
\label{app:phase0:budget}

\begin{figure}[htbp]
  \centering
  \includegraphics[width=0.95\linewidth]{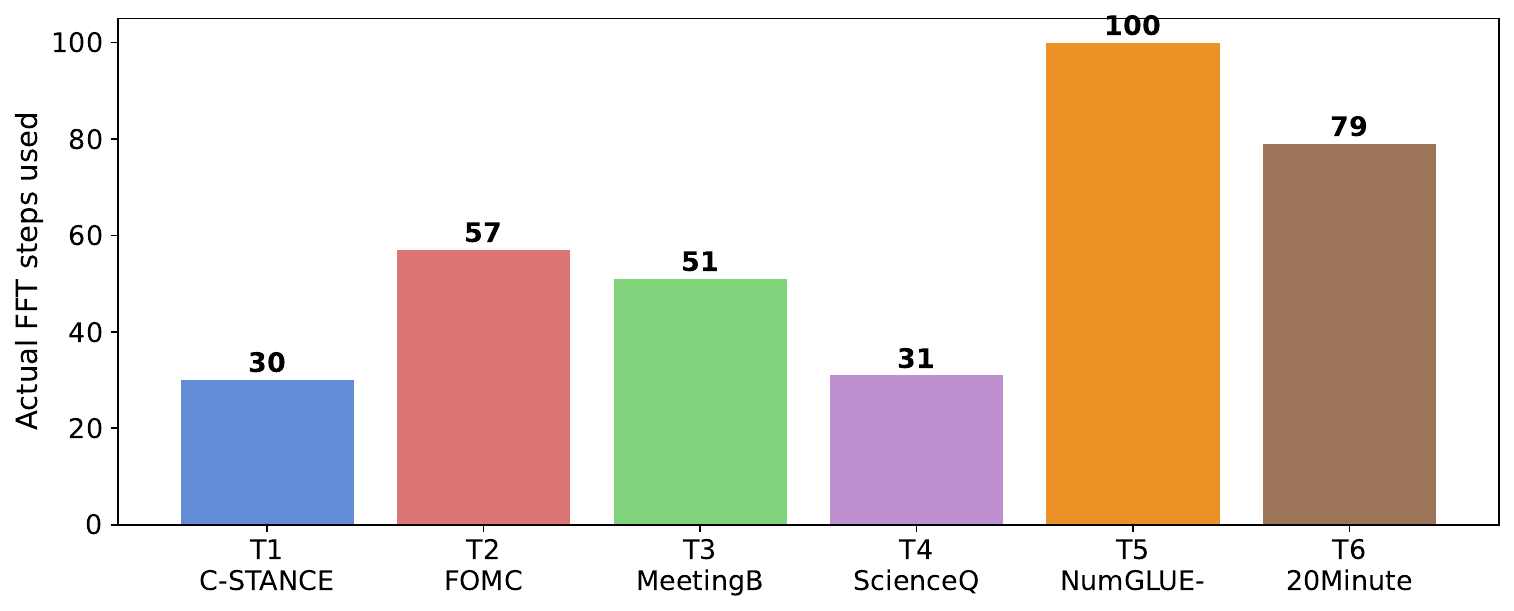}
  \caption{Actual FFT steps consumed per task under the dynamic stopping
    criterion for LLaMA-3 8B. Classification tasks such as C-STANCE and
    ScienceQA stabilise in under 35~steps; generation tasks such as
    NumGLUE-cm and 20Minuten require up to 100~steps, reflecting their
    more complex gradient landscapes.}
  \label{fig:fft_steps}
\end{figure}

The dynamic stopping criterion adapts the FFT budget to each task's
gradient complexity without any manual tuning. Tasks with concentrated
gradient signal converge quickly, while open-ended generation tasks
require longer stabilisation. This task-sensitivity directly validates
the adaptive design of Step~A.

\subsection{Cross-Task Block Overlap}
\label{app:phase0:overlap}

\begin{figure}[htbp]
  \centering
  \includegraphics[width=0.95\linewidth]{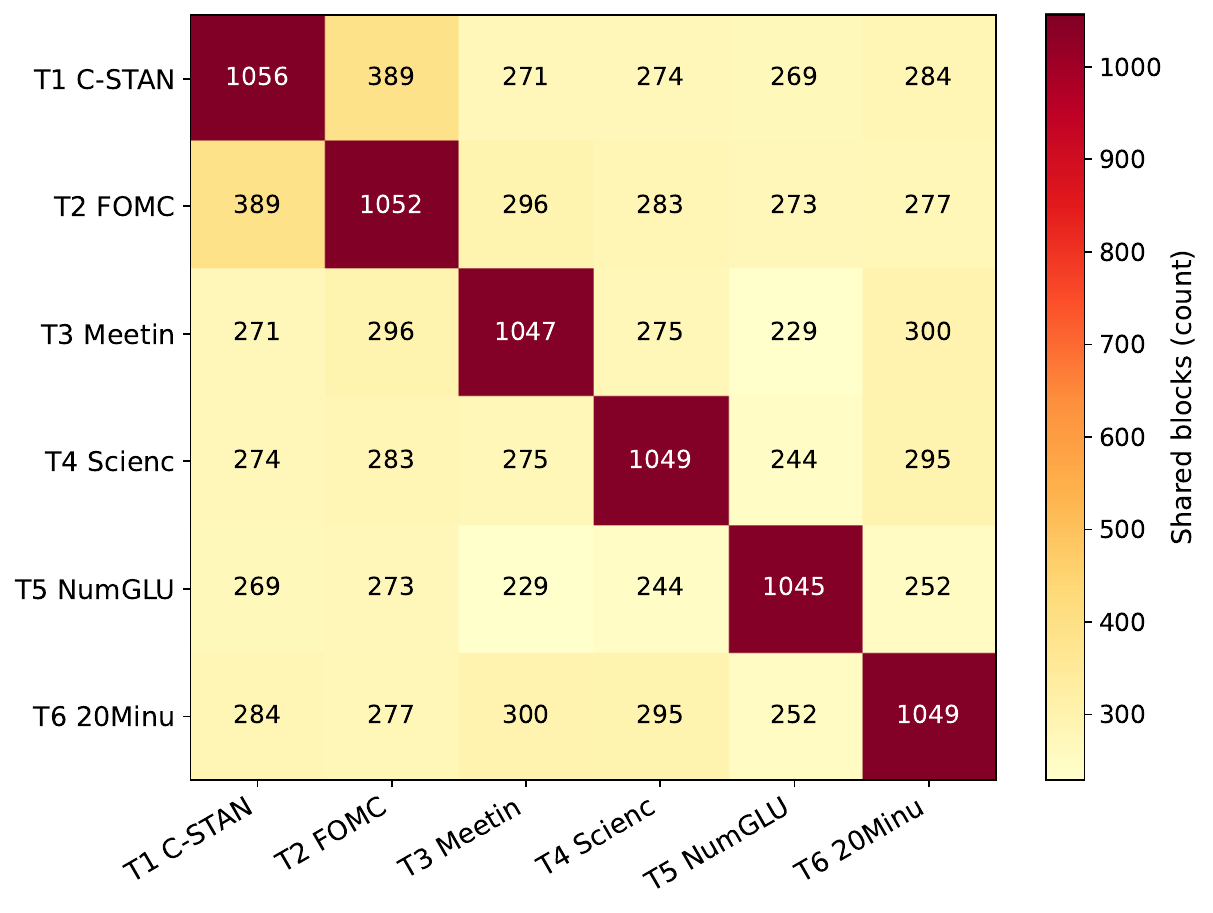}
  \caption{Cross-task block overlap matrix for LLaMA-2 7B. Diagonal
    entries report each task's total block budget, ranging from
    approximately 1{,}045 to 1{,}056; off-diagonal entries report the
    number of blocks shared between task pairs. Off-diagonal values span
    229--389 blocks, with the strongest overlap between C-STANCE and
    FOMC (389 blocks) and the weakest between MeetingBank and NumGLUE-cm
    (229 blocks). This heterogeneous but consistently partial overlap
    provides the raw material for shared experts in the SoS mechanism.}
  \label{fig:cross_task_overlap}
\end{figure}

\begin{figure}[htbp]
  \centering
  \includegraphics[width=\linewidth]{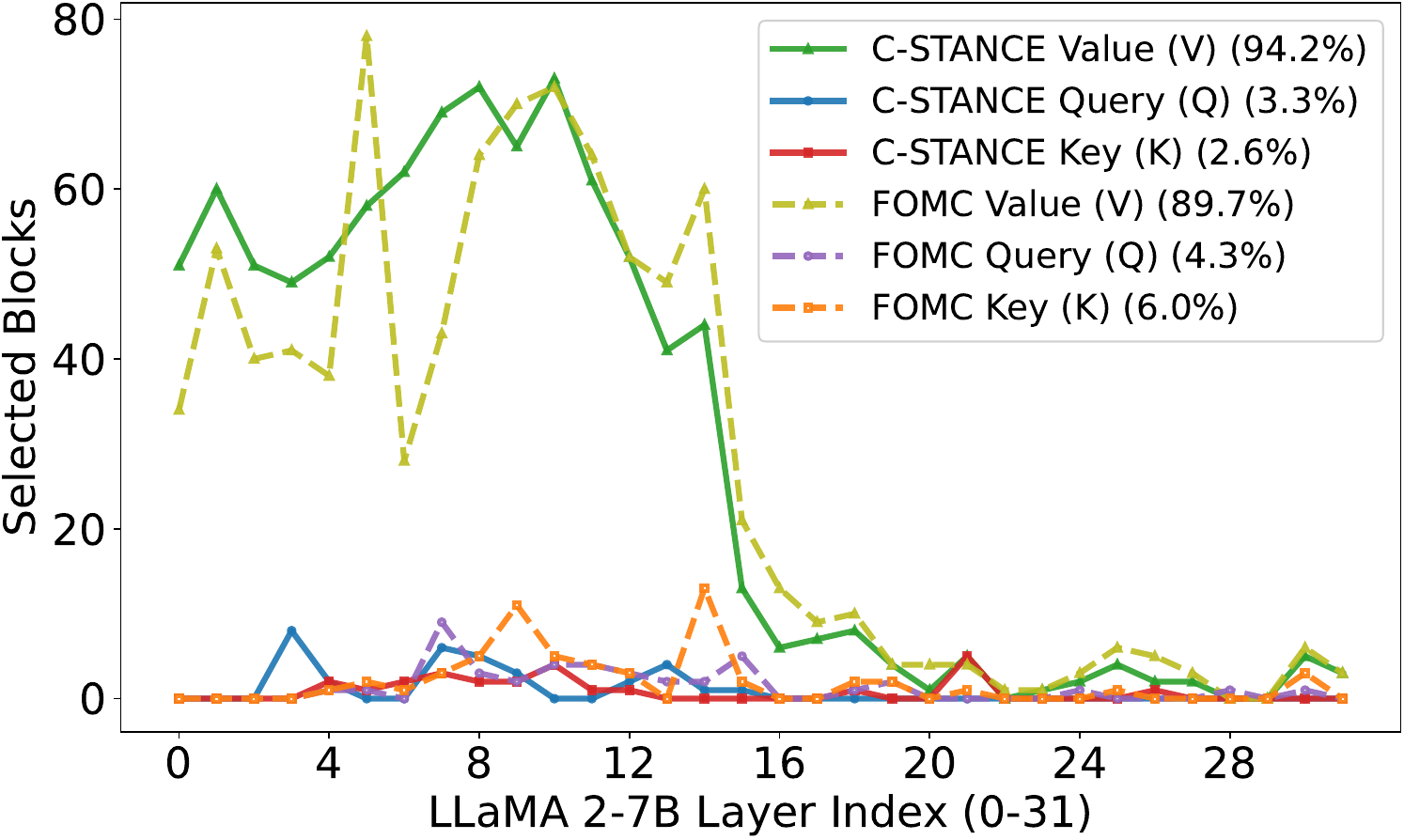}
  \caption{Layer-wise analysis shows $\sim$95\% of top-ranked gradients concentrate in the Value (V) projection, indicating that attention routing (Q, K) remains stable while content (V) requires adaptation.}
  \label{fig:gradient_analysis}
\end{figure}
No task pair shares more than 37\% of its blocks, confirming that each
task retains a substantially distinct gradient footprint. The partial
overlap is precisely what the SoS split exploits: blocks appearing in
multiple task selections are promoted to shared experts, while the
remainder become task-unique experts. The variation across pairs (229
to 389 shared blocks) further reflects the differing linguistic
proximity of the TRACE tasks stance detection and monetary-policy
classification share more representational structure than summarisation
and arithmetic reasoning.

\subsection{Sub-linear Parameter Growth}
\label{app:phase0:growth}

\begin{figure}[htbp]
  \centering
  \includegraphics[width=\linewidth]{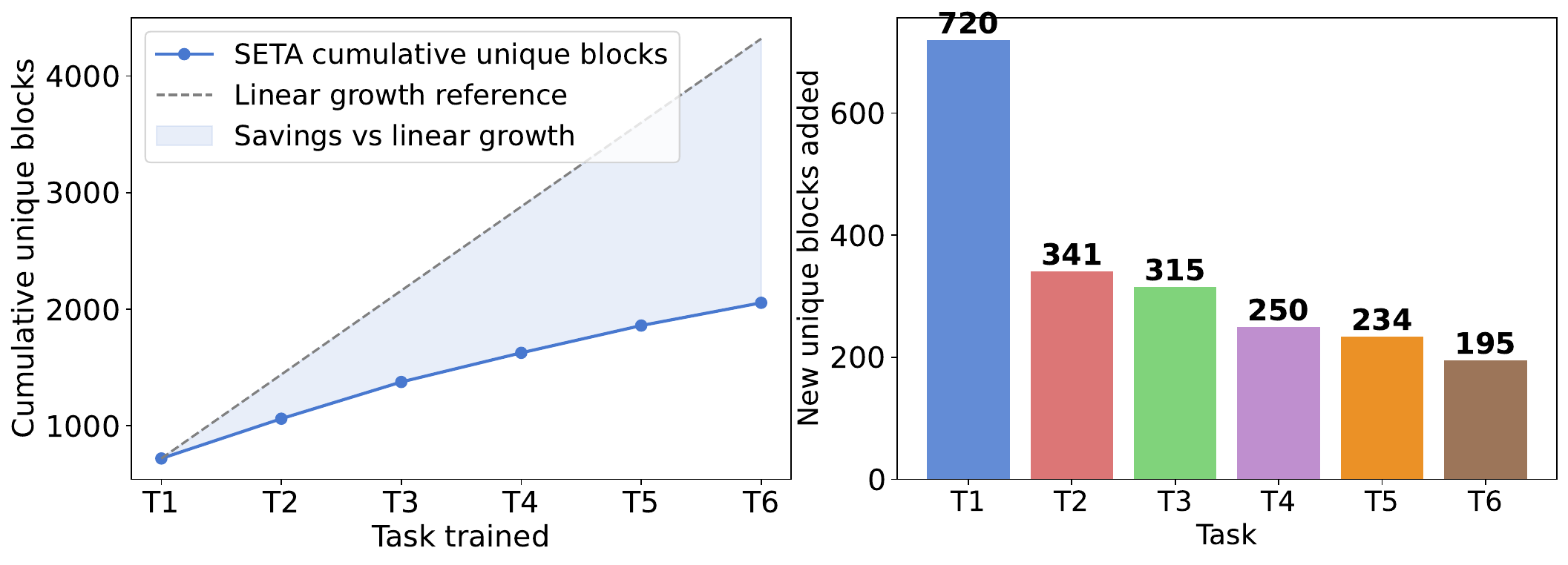}
  \caption{Cumulative unique blocks added across the task sequence for Qwen3-4B under 
the 13\% block budget. \emph{Left:} cumulative count against a linear growth 
reference (dashed); the shaded region quantifies capacity saved relative to a 
system that allocates fresh blocks per task. \emph{Right:} new unique blocks 
contributed by each task. Qwen3-4B adds 720 blocks for T1 but only 195 by T6, 
a 73\% reduction in per-task demand as more blocks are reused from earlier tasks.}
  \label{fig:cumulative_growth}
\end{figure}
Sub-linear growth is SETA's central efficiency claim. Each successive
task reuses a growing fraction of previously selected blocks, so the
total parameter footprint grows far slower than the number of tasks.
By T6, Qwen3-4B's cumulative unique block count of 2{,}055 is only 48\%
of what a non-reusing baseline would require at 4{,}320, a saving of
over 2{,}265 blocks. The per-task contribution drops monotonically from
720 at T1 to 195 at T6, confirming that later tasks operate
predominantly in the reuse regime.

\subsection{Novel Block Discovery per Layer}
\label{app:phase0:unique}
\begin{figure}[htbp]
  \centering
  \includegraphics[width=\linewidth]{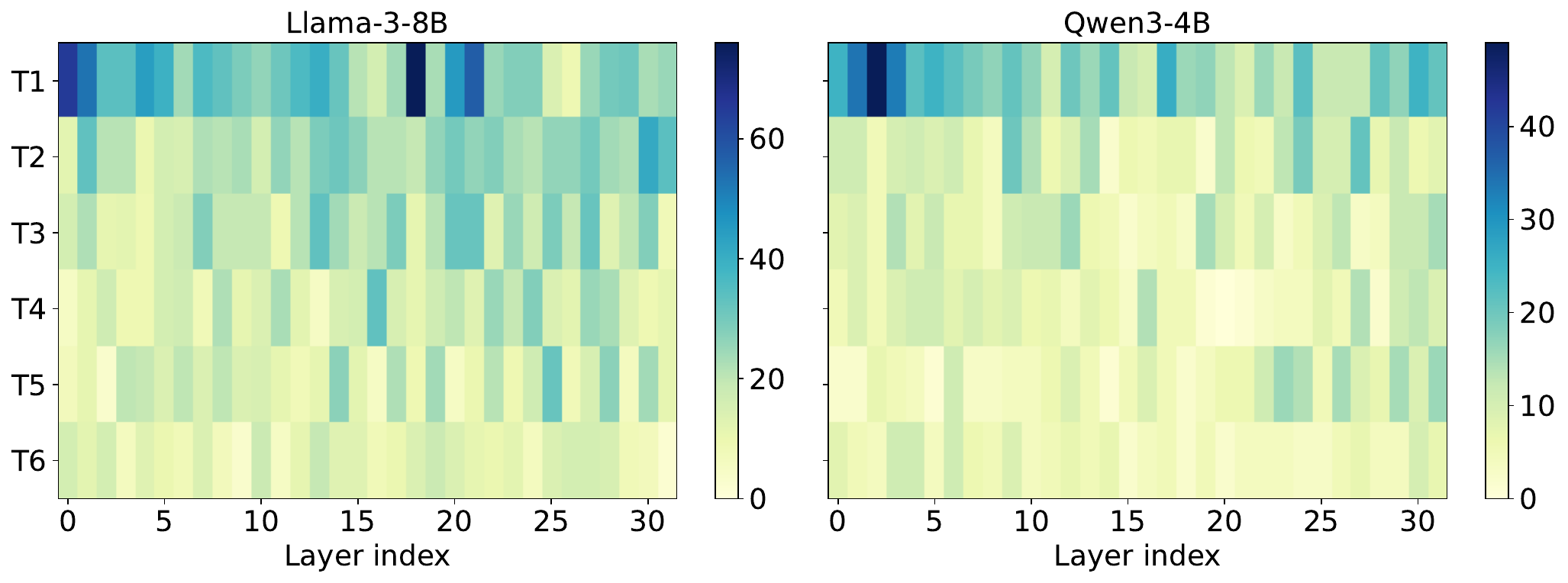}
  \caption{Unique blocks per transformer layer per task. A block is
    unique if no prior task selected it at the same layer position.
    Brighter cells indicate layers where a task introduced genuinely
    new block coverage. Later tasks T4--T6 show sparser brightness
    than T1--T2, reflecting increasing reuse as the sequence progresses.}
  \label{fig:unique_layer_heatmap}
\end{figure}

The progressive dimming of later task rows confirms that earlier tasks
have already claimed the most informative layer-block positions, and
later tasks operate increasingly in the reuse regime. This is the
layer-level evidence underlying the cumulative growth result in
Figure~\ref{fig:cumulative_growth}.

\subsection{Final Task Block Composition}
\label{app:phase0:t6}

\begin{figure}[htbp]
  \centering
  \includegraphics[width=\linewidth]{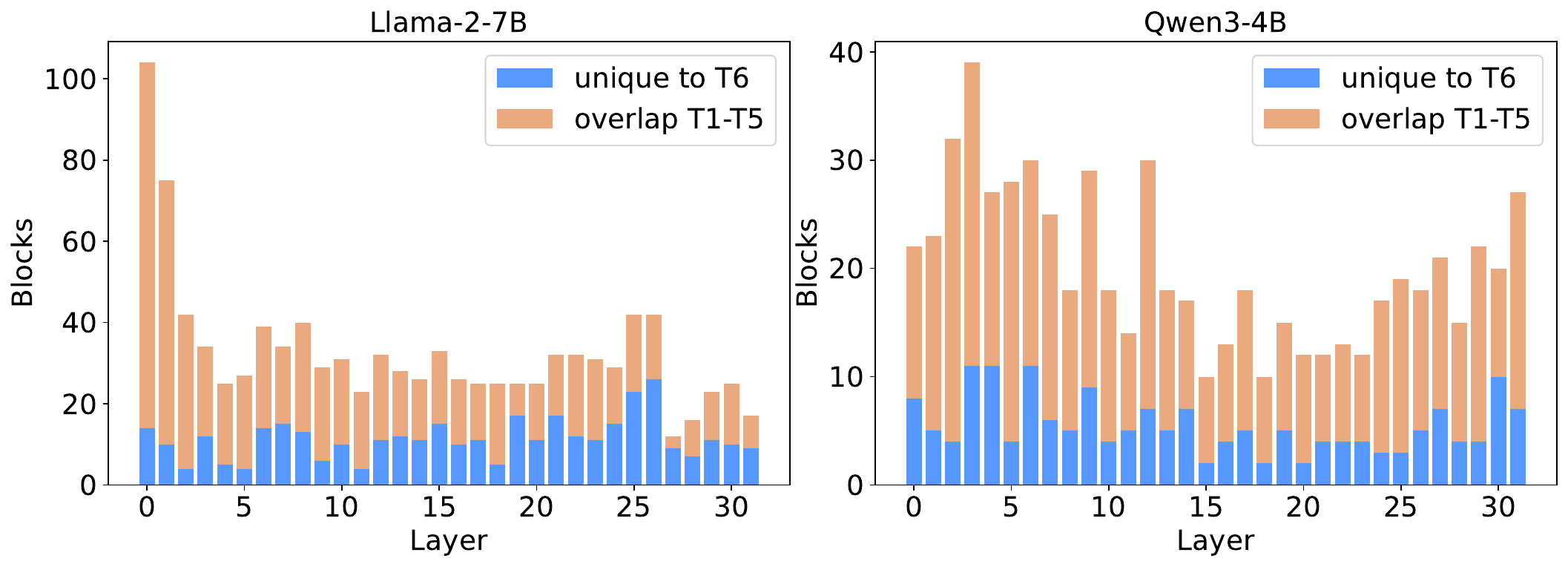}
  \caption{Per-layer block composition for the final task T6 (20Minuten)
    for LLaMA-2 7B (left, y-scale 0--100 blocks) and Qwen3-4B (right,
    y-scale 0--40 blocks). Blue bars show blocks unique to T6; orange
    bars show blocks shared with at least one prior task T1--T5. Both
    architectures show mixed composition across all layers, with no layer
    being entirely novel or entirely reused.}
  \label{fig:t6_overlap}
\end{figure}

Even at T6, after five prior tasks have collectively covered the
gradient landscape, the model still identifies genuine novel blocks in
most layers. This prevents expert saturation and ensures T6 receives a
dedicated unique expert. The layer-wise mixture of novel and reused
blocks is the direct trigger for the SoS split decision at T6.

\begin{figure}[htbp]
  \centering
  \includegraphics[width=\linewidth]{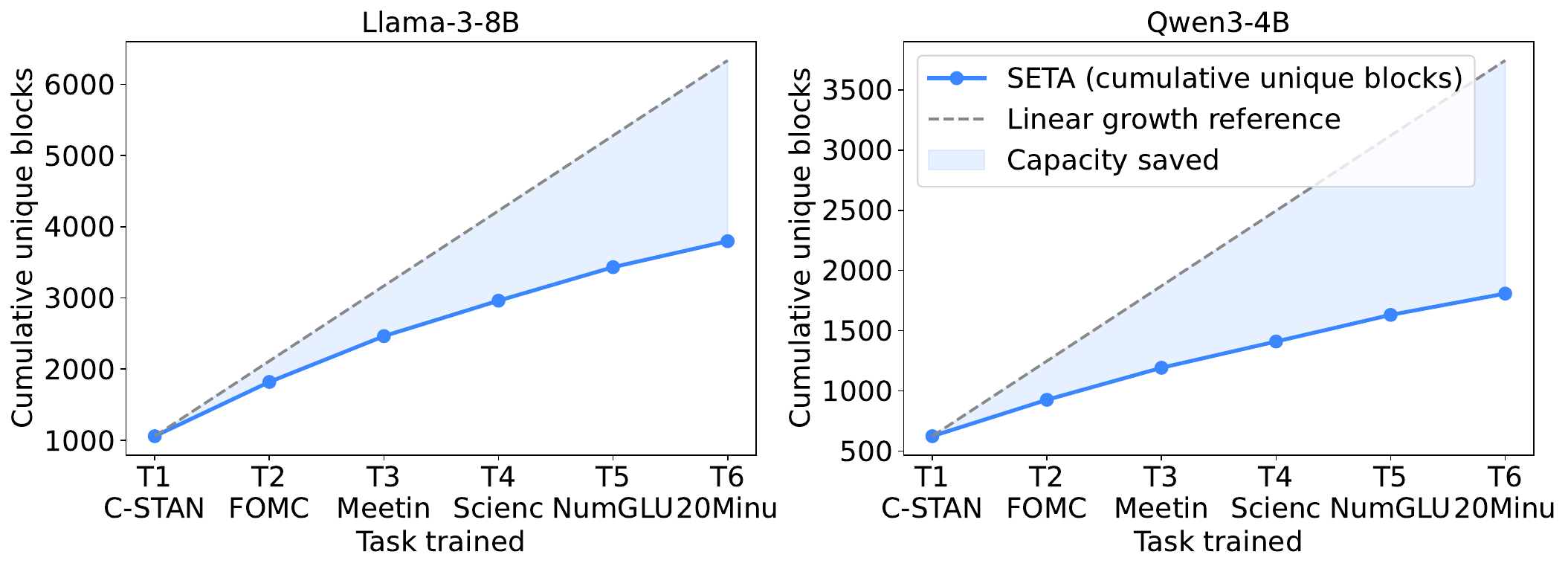}
  \caption{Cumulative unique block count across the task sequence for
    LLaMA-3 8B (left) and Qwen3-4B (right) against a linear growth
    reference (dashed). Both architectures exhibit consistent sub-linear
    growth, confirming model-agnostic efficiency through block reuse.
    Qwen3-4B converges toward the reuse regime slightly faster,
    consistent with its smaller attention dimensionality.}
  \label{fig:cumulative_models}
\end{figure}

\subsection{Expert Weight Analysis}
\label{app:expert_weight}

\begin{figure}[htbp]
    \centering
    \includegraphics[width=\linewidth]{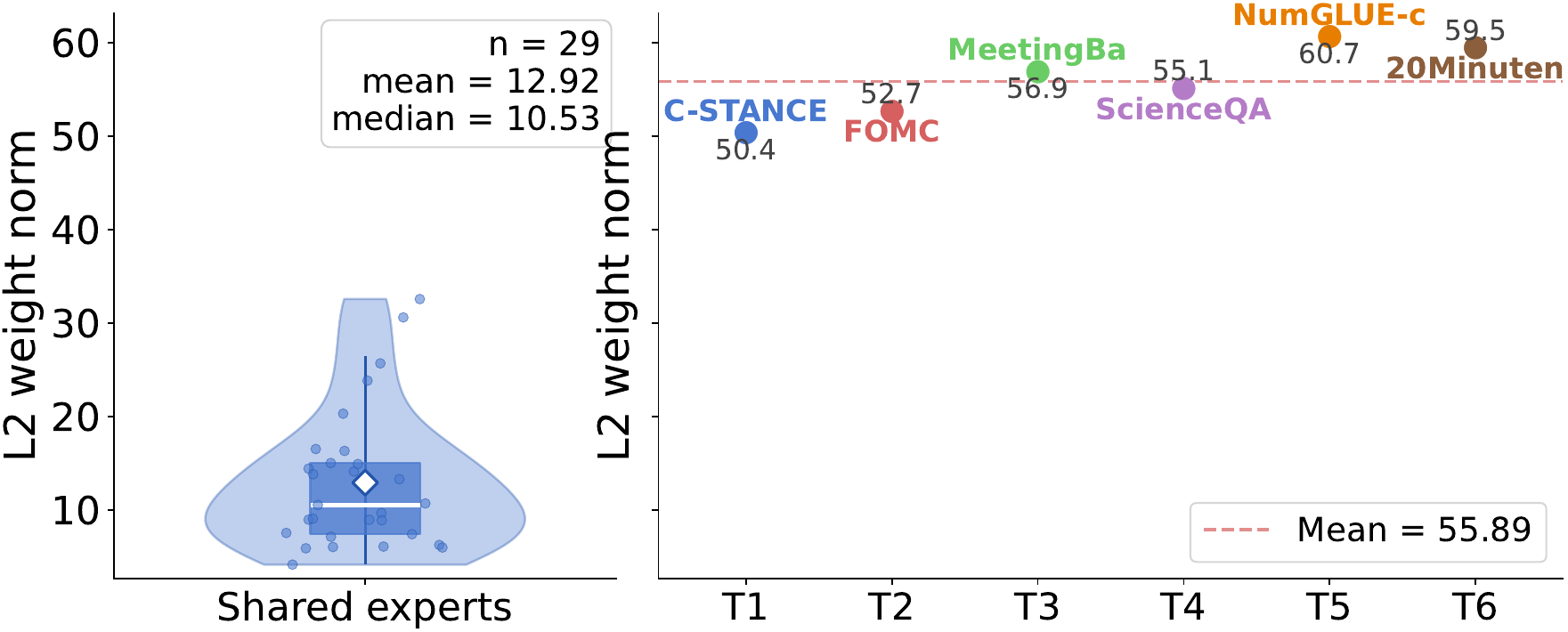}
    \caption{L2 weight norm of unique and shared experts in the final
    T6 SETA checkpoint for Qwen3-4B. Left: distribution of shared
    expert norms across $n=36$ experts; violin shows kernel density,
    inner box depicts the IQR with median as a white line, the diamond
    marks the mean, and jittered dots show individual expert norms.
    Right: each dot represents one unique expert trained for tasks
    T1--T6, coloured by task; the dashed line indicates the group mean.}
    \label{fig:weight_norm_shared_unique}
\end{figure}

\begin{figure}[htbp]
    \centering
    \includegraphics[width=\linewidth]{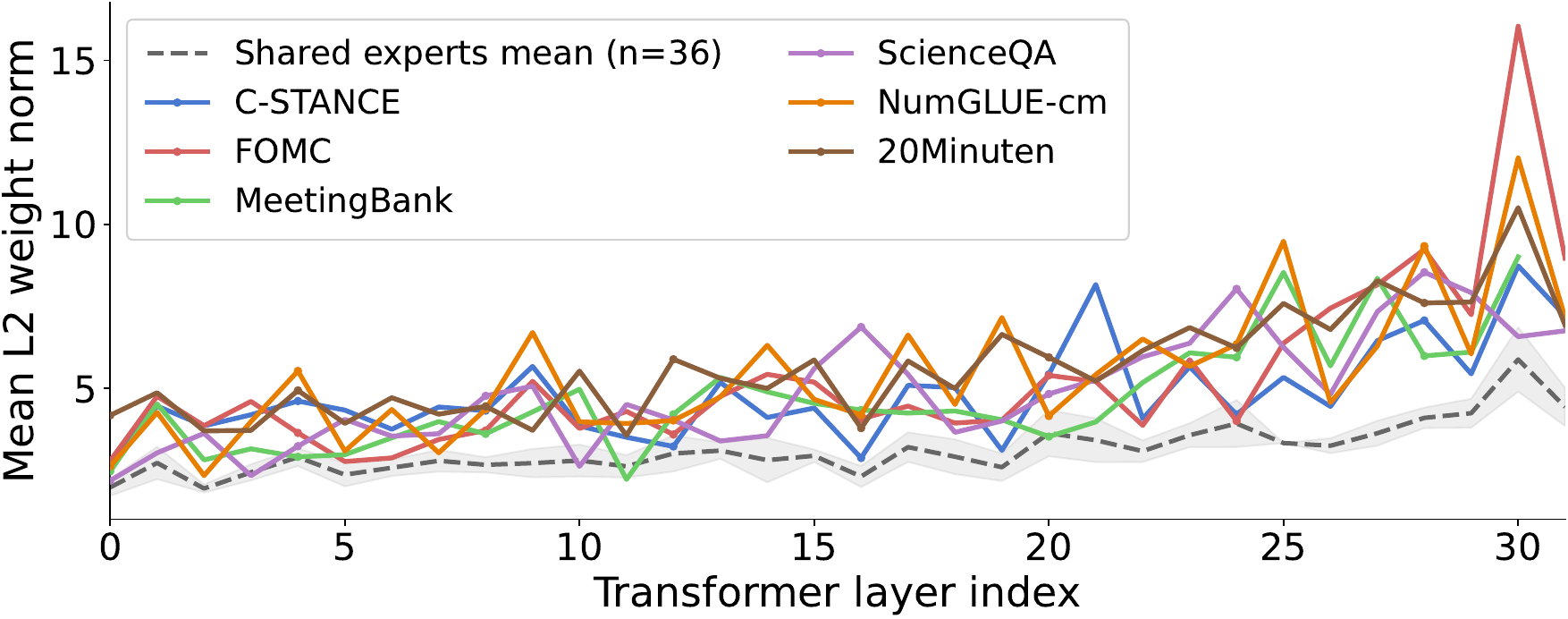}
    \caption{Mean L2 weight norm per transformer layer for each unique
    expert shown as coloured lines, one per task T1--T6, against the
    mean $\pm 1$ standard deviation of all shared experts shown as a
    dashed grey line with shaded band, for LLaMA-3 8B.}
    \label{fig:weight_norm_layerwise}
\end{figure}

Figures~\ref{fig:weight_norm_shared_unique}
and~\ref{fig:weight_norm_layerwise} examine the learned weight
magnitudes of unique and shared experts to verify that the SoS
architecture produces the intended structural separation between
task-specific and cross-task knowledge.

Figure~\ref{fig:weight_norm_shared_unique} reveals a clear structural
divergence between expert types. Unique experts attain substantially
higher L2 norms at a mean of approximately 30.8, compared to shared
experts at a mean of approximately 6.4, a gap of approximately
$5\times$. This confirms that unique experts undergo strong
task-specific gradient updates, while shared experts remain close to
the pre-trained base, acting as a regularised cross-task knowledge
store. The pattern is consistent with the SoS design: orthogonal
subspace isolation restricts gradient flow into prior unique experts,
concentrating plasticity in the current task subspace, while elastic
anchoring actively suppresses drift in shared experts.

\subsection{Expert Creation Threshold}
\label{app:threshold:ect_budget}

We analyse how the Expert Creation Threshold (ECT) and the block budget
fraction jointly govern the SoS split behaviour across architectures.

\begin{figure*}[htbp]
    \centering
    \includegraphics[width=\linewidth]{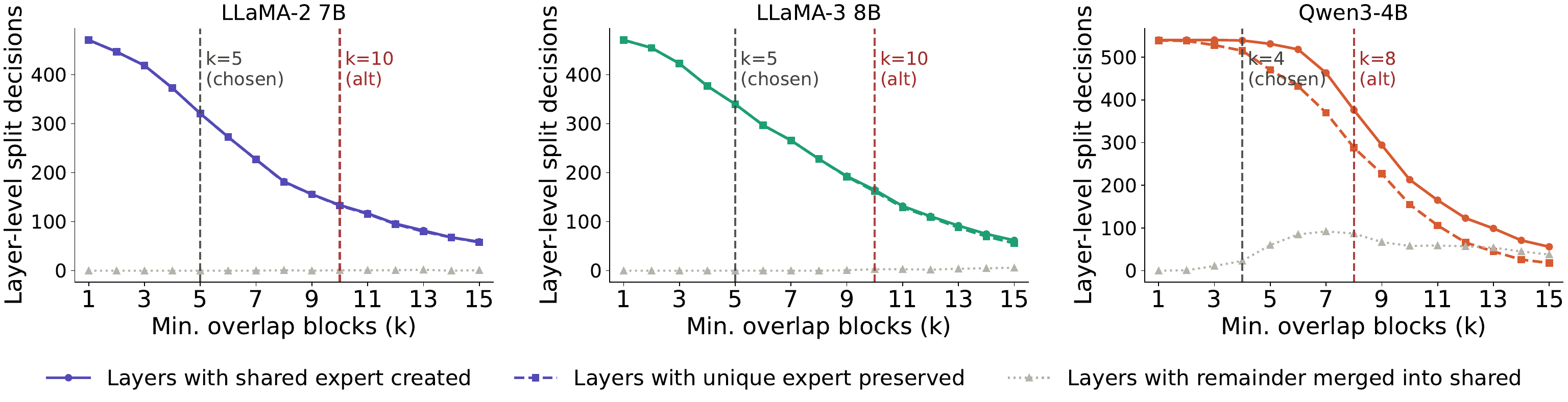}
    \caption{Layer-level split decisions as a function of ECT for
    LLaMA-2 7B, LLaMA-3 8B, and Qwen3-4B. Each curve counts layer
    decisions across all 15 task pairs: shared expert created, unique
    expert preserved, and remainder merged into shared. The black dashed
    line marks the chosen ECT; the red dashed line marks the alternative
    value for comparison.}
    \label{fig:trt3_diagonal_scan}
\end{figure*}

Figure~\ref{fig:trt3_diagonal_scan} shows that as the threshold
increases, shared expert creation decreases monotonically across all
architectures. For LLaMA-2 7B and LLaMA-3 8B, the curves decline
steeply around the chosen value, sitting at the elbow where small
increases yield large reductions in approved splits. For Qwen3-4B,
the decline occurs earlier due to its smaller per-layer block space,
making the chosen threshold the natural operating point. Across all
architectures, the absorbed remainder curve remains near zero,
indicating that tiny remainder fragments are rare in practice and the
Tiny Remainder Threshold has minimal absorptive effect.
\section{Baseline Descriptions}
\label{app:baseline}
To rigorously evaluate SETA, we benchmark against ten diverse methods spanning regularization, replay, and architecture-based Continual Learning (CL) strategies. All methods employ identical LoRA architectures and parameter budgets to ensure fair comparison.

\subsection{Baselines}
We compare SETA against PEFT-based baselines to ensure a fair evaluation. The methods include (1) Sequence Fine-tuning (Seq-LoRA), which adapts continuously without regularization; (2) Elastic Weight Consolidation (EWC)~\citep{kirkpatrick2017overcoming}, which restricts critical parameter changes via Fisher information. We also evaluate gradient-based constraints (3) GEM~\citep{lopez2017gradient}, (4) Progressive Prompt (PP)~\citep{razdaibiedina2023progressive} and (5) I-LoRA~\citep{li2025analyzing} included as a direct architectural comparator.
\paragraph{Traditional CL Strategies:}
\begin{itemize}
    \item Sequence Fine-tuning (Seq-LoRA): Represents naive adaptation, continuously tuning parameters on the task sequence without regularization or replay.
    \item Elastic Weight Consolidation (EWC)~\citep{kirkpatrick2017overcoming}: A regularization-based approach that utilizes the Fisher information matrix to penalize changes to parameters deemed critical for previous tasks.
    \item Gradient Episode Memory (GEM)~\citep{lopez2017gradient}: Maintains a gradient subspace of old tasks, projecting updates orthogonally to avoid interference.
\end{itemize}

\paragraph{Architecture and Prompt-Based Methods:}
\begin{itemize}
    \item I-LoRA~\citep{li2025analyzing}: A relevant comparative baseline that utilizes dual-memory LoRA adapters for sequential learning.
    \item Progressive Prompt (PP)~\citep{razdaibiedina2023progressive}: Sequentially learns task-specific soft prompts and concatenates them, isolating task knowledge in distinct prefix tokens.
\end{itemize}
\subsection{Prompt Design for Training and Inference}

To clarify, all models are utilized in a strictly generative text-to-text paradigm via causal language modeling. We do not append any task-specific classification heads to the architecture. Both training and inference are conducted using a standardized instruction-tuning prompt format. Every data instance across all tasks is uniformly structured with a system prefix, the task instruction including context, and a response trigger:

\begin{verbatim}
Below is an instruction that describes 
a task. 
Write a response that appropriately 
completes the request.

### Instruction:
{task_specific_instruction_and_input}

### Response:
{target_label_or_text}
\end{verbatim}

During evaluation, the model is provided the prompt up to the response trigger, and the output is generated autoregressively. Because we do not use a classification head, we map the raw generated text back to the discrete labels using dataset-specific parsing scripts. For classification tasks, the parsing functions extract the predicted option or numeric label from the generated string and compare it against the target ground truth to compute Accuracy or macro-F1 metrics. For generation tasks like MeetingBank, the raw generated text is directly evaluated against the reference summary using ROUGE metrics.
\section{Detailed Dataset Specifications}
\label{app:datasets}
To ensure a robust evaluation of both adaptation and forgetting, we utilized the following specific datasets categorized by their intended evaluation metrics.

\subsection{CL Benchmarks for LLMs}
We selected datasets to maximize domain specificity and task diversity across three primary dimensions. First, regarding \textbf{Domain Specificity}, we sourced datasets from five distinct verticals to simulate realistic and diverse downstream applications. These include ScienceQA~\citep{lu2022learn} for the educational domain, FOMC~\citep{shah2023trillion} for financial forecasting, and MeetingBank~\citep{hu2023meetingbank} for political discourse. Second, regarding \textbf{Multilinguality}, we addressed the hurdles of vocabulary variations by following established protocols and incorporating C-STANCE~\citep{zhao2023c} and 20Minuten~\citep{kew202320} to evaluate the bridging of linguistic gaps. Finally, for \textbf{Mathematical Reasoning}, we leveraged the NumGLUE dataset~\citep{mishra2022numglue} as a rigorous test bed for updating complex arithmetic and symbolic logical operations.

\section{Computational Overhead}
To formalize the computational analysis, let $B, S$, and $d$ denote the batch size, sequence length, and hidden dimension, respectively. For the adaptation and routing components, let $r$ represent the LoRA rank, $E$ the total number of experts, $K$ the active experts per token, and $L_{\text{SETA}}$ the number of routed layers.  The architectural design of SETA leverages this dynamic MoE routing to achieve superior expressive power with high computational efficiency. While i-LoRA is restricted to a static, monolithic adapter per task with a fixed $\mathcal{O}(rd)$ overhead, SETA utilizes conditional computation to activate only a sparse subset ($K \ll E$) of task-specific experts. This routing mechanism requires a minimal gating overhead of $\mathcal{O}(dE)$, yet it enables the model to dynamically specialize at the token level with an execution cost of only $\mathcal{O}(Krd)$. By decoupling the total parameter capacity from the per-token compute, SETA provides an adaptive framework that scales far more effectively than i-LoRA's rigid structure, allowing for high-fidelity task navigation through efficient, on-demand parameter fetching.
\section{SETA Algorithm}
Algorithm~\ref{alg:seta_algo} details the SETA training procedure.
For each task $t$, Line~3 identifies the active sparse subspace
$\mathcal{P}_t$ via an adaptive gradient-based pipeline. For $t>1$,
the Split-on-Share mechanism in Lines~7--11 decomposes this subspace
layer-wise: overlapping parameters merge into the shared expert
$E_{\text{s}}$, disjoint parameters form the frozen unique expert
$E_{\text{u}}^{(t-1)}$, and novel indices initialize $E_{\text{u}}^{(t)}$.
Prior experts are frozen in Line~13, the gating network expanded in
Line~14, and the model optimized in Line~15 over
$\mathcal{L}_{\text{task}} + \mathcal{L}_{\text{reg}} +
\mathcal{L}_{\text{gate}}$. Line~17 snapshots shared expert weights
before the next task begins.
\label{app:algorithm}
\begin{algorithm}[h]
\caption{SETA: Sparse Subspace-to-Expert Training Procedure}
\label{alg:seta_algo}
\begin{algorithmic}[1]
\REQUIRE Pre-trained $W_0$, Task Sequence $\mathcal{T} = \{T_1, \dots, T_k\}$,
         Thresholds $\tau_{\text{ect}}, \tau_{\text{trt}}$,
         Regularization coefficient $\lambda_{\text{base}}$
\STATE \textbf{Initialize:} $\mathcal{E} \leftarrow \emptyset$,
       $W_g \leftarrow W_0$
\FOR{$t = 1$ to $k$}
    \STATE \textbf{Subspace Selection:} Identify active sparse blocks
           $\mathcal{P}_t$ via adaptive gradient pipeline.
    \IF{$t=1$}
        \STATE Construct initial unique expert $E_{\text{u}}^{(1)}$
               on $\mathcal{P}_1$.
    \ELSE
        \FOR{each layer $l \in L$}
            \STATE \textbf{SoS Decomposition:} Compute $\mathcal{I}_l$
                   and $\mathcal{R}_l$ via layer-wise topological filters
                   (Eq.~\ref{eq:sos_Filtering_Small_Overlaps},~\ref{eq:sos_Merging_Tiny_Fragments}).
            \STATE \textbf{Expert Allocation:} $E_{\text{s}} \leftarrow
                   \mathcal{I}_l$ via weight inheritance;
                   freeze $E_{\text{u}}^{(t-1)} \leftarrow \mathcal{R}_l$.
            \STATE \textbf{New Expert:} Initialize $E_{\text{u}}^{(t)}$
                   on $\mathcal{P}_t \setminus (\mathcal{I}_l \cup \mathcal{R}_l)$.
        \ENDFOR
    \ENDIF
    \STATE \textbf{Freeze History:} $\nabla E_{\text{u}}^{(k)} = 0$
           for all $k < t$.
    \STATE \textbf{Gating Expansion:} Resize $W_g$ to $|\mathcal{E}|$;
           preserve logit invariance.
    \STATE \textbf{Optimize:} Minimize $\mathcal{L}_{\text{task}} +
           \mathcal{L}_{\text{reg}} + \mathcal{L}_{\text{gate}}$
           over $E_{\text{u}}^{(t)}$ and $E_{\text{s}}$
           (Eq.~\ref{equ:seta_train_object}).
    \STATE \textbf{Anchor Update:} $\hat{W}_i \leftarrow W_i$ for all
           shared experts.
\ENDFOR
\STATE \textbf{Inference:} Task-agnostic retrieval via $W_g$
       (Eq.~\ref{eq:task_agnostic_pred}).
\end{algorithmic}
\end{algorithm}

\section{Detailed Per-Task Results Across Models and Baselines}
\begin{table*}[t]
\centering
\caption{Task-wise inference accuracy of SeqLoRA on LLaMA-2 7B. Diagonal 
entries reflect plasticity; off-diagonal decay reflects forgetting.}
\label{tab:seqlora_llama2_7b}
\footnotesize
\begin{tabular}{lcccccc}
\toprule
\textbf{SeqLoRA} & \textbf{C-STANCE} & \textbf{FOMC} & \textbf{MeetingBank} & \textbf{ScienceQA} & \textbf{NumGLUE-cm} & \textbf{20Minuten} \\
\midrule
C-STANCE    & 41.8 &      &      &      &      &      \\
FOMC        & 21.8 & 60.3 &      &      &      &      \\
MeetingBank & 19.4 & 49.4 & 24.7 &      &      &      \\
ScienceQA   & 32.0 & 26.2 & 14.8 & 36.8 &      &      \\
NumGLUE-cm  & 25.6 & 20.4 &  5.7 & 24.4 & 33.3 &      \\
20Minuten   &  0.2 &  1.2 &  8.7 & 15.8 & 14.8 & 41.4 \\
\bottomrule
\end{tabular}
\end{table*}

% Table 3: EWC LLaMA-2 7B
\begin{table*}[htbp]
\centering
\caption{Task-wise inference accuracy of EWC on LLaMA-2 7B. Diagonal 
entries reflect plasticity; off-diagonal decay reflects forgetting.}
\label{tab:ewc_llama2_7b}
\footnotesize
\begin{tabular}{lcccccc}
\toprule
\textbf{EWC} & \textbf{C-STANCE} & \textbf{FOMC} & \textbf{MeetingBank} & \textbf{ScienceQA} & \textbf{NumGLUE-cm} & \textbf{20Minuten} \\
\midrule
C-STANCE    & 42.2 &      &      &      &      &      \\
FOMC        & 35.6 & 70.6 &      &      &      &      \\
MeetingBank & 33.8 & 48.4 & 25.6 &      &      &      \\
ScienceQA   & 37.4 & 38.1 & 14.3 & 64.2 &      &      \\
NumGLUE-cm  & 30.8 & 25.0 &  6.8 & 36.8 & 28.4 &      \\
20Minuten   & 19.0 & 24.8 & 15.2 & 42.4 & 14.8 & 41.5 \\
\bottomrule
\end{tabular}
\end{table*}

% Table 4: GEM LLaMA-2 7B
\begin{table*}[htbp]
\centering
\caption{Task-wise inference accuracy of GEM on LLaMA-2 7B. Diagonal 
entries reflect plasticity; off-diagonal decay reflects forgetting.}
\label{tab:gem_llama2_7b}
\footnotesize
\begin{tabular}{lcccccc}
\toprule
\textbf{GEM} & \textbf{C-STANCE} & \textbf{FOMC} & \textbf{MeetingBank} & \textbf{ScienceQA} & \textbf{NumGLUE-cm} & \textbf{20Minuten} \\
\midrule
C-STANCE    & 38.8 &      &      &      &      &      \\
FOMC        & 31.4 & 61.5 &      &      &      &      \\
MeetingBank & 17.4 & 44.4 & 23.3 &      &      &      \\
ScienceQA   & 31.8 & 26.0 & 11.5 & 40.4 &      &      \\
NumGLUE-cm  &  5.0 &  0.6 &  7.4 & 18.6 & 32.1 &      \\
20Minuten   &  5.2 &  7.1 & 12.0 & 24.0 & 17.3 & 40.9 \\
\bottomrule
\end{tabular}
\end{table*}

\begin{table*}[htbp]
\centering
\caption{Task-wise inference accuracy of I-LoRA on Qwen3-4B. Diagonal 
entries reflect plasticity; off-diagonal decay reflects forgetting.}

\label{tab:ilora_pertask}
\footnotesize
\begin{tabular}{lcccccc}
\toprule
 & C-STANCE & FOMC & MeetingBank & ScienceQA & NumGLUE-cm & 20Minuten \\
\midrule
C-STANCE    & 59.0 &      &      &      &      &      \\
FOMC        & 50.6 & 66.3 &      &      &      &      \\
MeetingBank & 15.8 & 59.5 & 19.0 &      &      &      \\
ScienceQA   & 33.0 & 45.8 & 16.4 & 82.4 &      &      \\
NumGLUE-cm  & 19.8 & 45.4 & 18.4 & 80.6 & 40.7 &      \\
20Minuten   & 27.2 & 38.5 & 16.0 & 71.0 & 37.0 & 39.2 \\
\bottomrule
\end{tabular}
\end{table*}
\begin{table*}[htbp]
\centering
\caption{Task-wise inference accuracy of EWC on Qwen3-4B. Diagonal 
entries reflect plasticity; off-diagonal decay reflects forgetting.}
\label{tab:ewc_qwen3_4b_pertask}
\footnotesize
\begin{tabular}{lcccccc}
\toprule
 & C-STANCE & FOMC & MeetingBank & ScienceQA & NumGLUE-cm & 20Minuten \\
\midrule
C-STANCE    & 59.8 &      &      &      &      &      \\
FOMC        & 50.4 & 67.3 &      &      &      &      \\
MeetingBank &  0.2 & 42.3 & 19.2 &      &      &      \\
ScienceQA   &  5.2 & 36.1 & 22.0 & 84.2 &      &      \\
NumGLUE-cm  &  5.8 & 52.2 & 17.0 & 75.2 & 60.5 &      \\
20Minuten   &  0.2 &  0.6 &  4.2 &  6.4 & 46.9 & 38.6 \\
\bottomrule
\end{tabular}
\end{table*}
\begin{table*}[t]
\centering
\caption{Task-wise inference accuracy of SeqLoRA on Qwen3-4B. Diagonal 
entries reflect plasticity; off-diagonal decay reflects forgetting.}

\label{tab:seq_qwen3_4b_pertask}
\footnotesize
\begin{tabular}{lcccccc}
\toprule
 & C-STANCE & FOMC & MeetingBank & ScienceQA & NumGLUE-cm & 20Minuten \\
\midrule
C-STANCE    & 60.0 &      &      &      &      &      \\
FOMC        & 49.7 & 67.5 &      &      &      &      \\
MeetingBank &  0.0 & 46.0 & 19.8 &      &      &      \\
ScienceQA   &  0.0 & 27.3 & 22.9 & 85.5 &      &      \\
NumGLUE-cm  &  0.0 & 19.8 & 12.1 & 75.2 & 56.8 &      \\
20Minuten   &  0.2 & 21.5 &  5.2 & 33.3 & 51.9 & 38.9 \\
\bottomrule
\end{tabular}
\end{table*}

\begin{table*}[t]
\centering
\caption{Task-wise inference accuracy of GEM on Qwen3-4B. Diagonal 
entries reflect plasticity; off-diagonal decay reflects forgetting.}
\label{tab:gem_qwen3_4b}
\footnotesize
\begin{tabular}{lcccccc}
\toprule
 & C-STANCE & FOMC & MeetingBank & ScienceQA & NumGLUE-cm & 20Minuten \\
\midrule
C-STANCE    & 60.3 &      &      &      &      &      \\
FOMC        & 48.7 & 68.5 &      &      &      &      \\
MeetingBank &  0.0 & 47.2 & 18.8 &      &      &      \\
ScienceQA   &  8.0 & 42.0 & 19.6 & 83.8 &      &      \\
NumGLUE-cm  &  2.8 & 51.2 & 11.9 & 80.5 & 49.4 &      \\
20Minuten   &  0.5 & 13.5 &  6.5 & 67.7 & 29.6 & 38.6 \\
\bottomrule
\end{tabular}
\end{table*}
\end{document}